\documentclass[10pt,final, twocolumn]{IEEEtran}
\usepackage{amsmath,graphicx}
\usepackage{amssymb}
\usepackage{acronym}

\usepackage[belowskip=-10pt,aboveskip=1.8pt]{caption}
\usepackage{standalone}
\usepackage{subcaption}
\usepackage{tikz,cite}
\usetikzlibrary{arrows.meta, shapes.geometric, positioning}
\usepackage{url,enumitem}
\usepackage{verbatim}
\usepackage[bookmarks,colorlinks]{hyperref}
\usepackage{soul, xcolor}
\usepackage{adjustbox}
\usepackage{booktabs}
\usepackage{lscape}
\usepackage{makecell}

\usepackage[ruled,linesnumbered,vlined]{algorithm2e}
\SetKwInput{KwData}{\textbf{Init}} 

\newcommand{\myVec}[1]{{\boldsymbol{#1}}}
\newcommand{\myMat}[1]{{\boldsymbol{#1}}}
\newcommand{\mySet}[1]{\mathcal{#1}}

\acrodef{snr}[SNR]{signal-to-noise ratio}
\acrodef{bs}[BS]{base station} 
\acrodef{cpu}[CPU]{centralized processing unit} 
\acrodef{mimo}[MIMO]{multiple-input multiple-output}
\acrodef{awgn}[AWGN]{additive white Gaussian noise} 
\acrodef{dnn}[DNN]{deep neural network} 
\acrodef{ml}[ML]{machine learning} 
\acrodef{mse}[MSE]{mean-squared error}
\acrodef{fft}[FFT]{fast fourier transform}
\acrodef{dct}[DCT]{discrete cosine transform}
\acrodef{iot}[IOT]{Internet of Things}
\acrodef{rmse}[RMSE]{root mean squared error}
\acrodef{rmspe}[RMSPE]{root mean squared periodic error}
\acrodef{mmse}[MMSE]{{minimum mean-squared error}}
\acrodef{lmmse}[LMMSE]{{linear} MMSE}
\acrodef{mle}[MLE]{maximum likelihood estimation}
\acrodef{admm}[ADMM]{alternating direction method of multipliers}
\acrodef{dadmm}[D-ADMM]{distributed alternating direction method of multipliers}
\acrodef{sar}[SAR]{successive approximation register}
\acrodef{adc}[ADC]{analog-to-digital converter} 
\acrodef{dac}[DAC]{digital-to-analog converter} 
\acrodef{msb}[MSB]{most significant bit}
\acrodef{lsb}[LSB]{least significant bit}
\acrodef{sh}[S\&H]{sample-and-hold}
\acrodef{sgd}[SGD]{stochastic gradient descent}
\acrodef{opa}[OpAmps]{operational amplifiers}
\acrodef{cnn}[CNN]{convolutional neural networks}

\usetikzlibrary{trees,arrows}

\setstcolor{blue}

\title{ Learning Task-Based Trainable Neuromorphic ADCs via Power-Aware Distillation
}
\author{
\IEEEauthorblockN{Tal Vol, Loai Danial, and Nir Shlezinger
\thanks{
Parts of this work were presented in the IEEE International Conference on Acoustics, Speech, and Signal Processing (ICASSP) 2024 as the paper \cite{vol2024power}. 
T. Vol and N. Shlezinger with the School of ECE, Ben-Gurion University of the Negev, Israel (e-mail: talvoldigital@gmail.com; nirshl@bgu.ac.il). 
L. Danial is with Mobileye Vision Technologies LTD, Haifa, Israel (e-mail:danial.loai@gmail.com).
}}}

\begin{document}
%\ninept

\maketitle
%
%%%%%%%%%%%%%%%%
%%% Abstract %%%
%%%%%%%%%%%%%%%%
%
\begin{abstract}
The ability to process signals in digital form depends on \acp{adc}. Traditionally, \acp{adc} are designed to ensure that the digital representation closely matches the analog signal. However, recent studies have shown that significant power and memory savings can be achieved through {\em task-based acquisition}, where the acquisition process is tailored to the  downstream processing task. An emerging technology for task-based acquisition involves the use of memristors, which are considered key enablers for neuromorphic computing. Memristors can implement \acp{adc} with tunable mappings, allowing adaptation to specific system tasks or power constraints.
%Analog-to-digital converters (ADCs) are essential components in many acquisition systems. While ADCs are integral to these systems, they often pose challenges due to their significant power consumption. Traditionally, the performance of an ADC is judged by the accuracy of its quantized results compared to the original signal. However, in task-based systems, the focus is not always on the precise quantized value but rather on performing specific tasks, such as classification or estimation, with minimal power consumption and high accuracy.
In this work, we study task-based acquisition for a generic classification task using memristive \acp{adc}. We consider the unique characteristics of this such neuromorphic \acp{adc}, including their power consumption and noisy read-write behavior, and propose a physically compliant model based on resistive \acl{sar} \acp{adc} integrated with memristor components, enabling the adjustment of quantization regions.
To optimize performance, we introduce a data-driven algorithm that jointly tunes task-based memristive \acp{adc} alongside both digital and analog processing. Our design addresses the inherent stochasticity of memristors through power-aware distillation, complemented by a specialized learning algorithm that adapts to their unique analog-to-digital mapping. The proposed approach is shown to enhance accuracy by up to 27\% and reduce power consumption by up to 66\% compared to uniform \acp{adc}. Even under noisy conditions, our method achieves substantial gains, with accuracy improvements of up to 19\% and power reductions of up to 57\%. These results highlight the effectiveness of our power-aware neuromorphic \acp{adc} in improving system performance across diverse tasks.
%Specifically, we observe a substantial enhancement in classification accuracy, achieving a factor of up to 1.5 times improvement. Most notably, our full optimized network yields a reduction in power consumption by as much as 75\%, substantiating its utility for energy-efficient computing applications.
\end{abstract}

\acresetall

%----------------------------------------------------------------------------------------
%	INTRODUCTION
%----------------------------------------------------------------------------------------
%\vspace{-0.2cm}
\section{Introduction}
\label{sec:intro}
%\vspace{-0.15cm}
% % Background 
\Acp{adc} play a key role in digital signal processing systems. Various applications, ranging from wireless communications and radar to medical imaging, involve simultaneous acquisition of multiple signals at high rates~\cite{chen2024information}. However, when acquiring multiple analog signals, particularly at high rates, the power consumption of conventional \acp{adc} can become a critical bottleneck, limiting the overall efficiency and performance of the system. Therefore, there is a growing need for low-power,
energy-efficient digital acquisition designs that can be translated into concrete \ac{adc} implementations.

Existing strategies to achieve energy-efficient digital acquisition are generally divided into {\em hardware} and {\em algorithmic} methods. Hardware-oriented solutions focus on developing \ac{adc} implementations, such as \ac{sar} \acp{adc}, sigma-delta \acp{adc}, and flash \acp{adc}, that can offer different trade-offs between power consumption, rate, and resolution~\cite{ADCSurvey,dalmia2020analog}. An emerging technology in this field is based on {\em memristors}, which offer a potential pathway toward adaptive and power-efficient \acp{adc}~\cite{FlashMem, si2023memristor, ADCMem}. Memristors are considered as an enabler technology for realizing neuromorphic computing~\cite{mead1990neuromorphic,yang2022research, aguirre2024hardware} and were recently shown to support implementation of innovative \ac{adc} architectures~\cite{danial2018breaking, PipelineADC,si2023memristor,guo2015modeling}. Such memristive \acp{adc} provide adaptivity and controllability, which comes at the cost of some inherent level of noise induced when dynamically writing and reading its conductance~\cite{LoaiNoise,Joshi2020, ErrorEst, Dupraz}. In the context of \ac{adc} design, the adaptivity of memristive \acp{adc} was shown to allow to optimize power efficiency while maintaining uniform signal mappings~\cite{danial2018breaking}. As hardware-oriented designs focus on the \ac{adc} circuitry, these approaches generally aim at realizing standard uniform \ac{adc} mappings in a manner which is power- and cost-efficient. 

%While these techniques emphasize converting signals into digital form for reconstruction, practical applications often aim to extract specific information from the signal. Incorporating this element directly into hardware components poses significant challenges, highlighting the need for innovative approaches. Modern research increasingly focuses on algorithmic strategies for acquisition, aiming to obtain concise representations that facilitate task-specific information extraction, further enhancing \ac{adc} systems' efficiency and applicability.

In parallel to \ac{adc} hardware advancements, various algorithmic developments were proposed to facilitate efficient signal acquisition. Such algorithmic methods aim at exploiting inherent structures in the signal, such as sparsity in a given domain~\cite{kipnis2018analog}, or knowledge that the signal is acquired for some lower dimensional downstream task~\cite{zou2022goal,wen2023task}, in order to mitigate the effect of the distortion induced by low power (and low resolution/rate) \acp{adc}. In particular, the exploitation of downstream tasks for efficient digital representation is considered to be vital for the emerging paradigms of goal-oriented and semantic communications~\cite{gunduz2022beyond,shi2023task,getu2024survey,lu2023semantics}. 

Algorithmic methods exploiting downstream tasks, known as {\em task-based quantization}, were analytically derived for low-resolution acquisition 
of signals for tasks modeled as estimating linear~\cite{shlezinger2018hardware,shlezinger2018asymptotic,neuhaus2020task} and quadratic functions~\cite{Salamtian19task,bernardo2023design}. For more complex tasks, data-driven techniques leveraging deep learning capabilities have shown significant improvements in designing acquisition schemes~\cite{mashhadi2020distributed, torfason2018towards, shlezinger2021deep, shao2021learning, shlezinger2022deep}. It was shown that for both analytically designed task-based acquisition systems~\cite{bernardo2023design}, as well as for data-driven designs~\cite{shlezinger2022deep}, performance notably improves by setting non-uniform \ac{adc} mappings along with overall processing.
However, existing algorithmic methods typically consider cost only in terms of the number of bits and do not account for \ac{adc} hardware, modeling it as a controllable mapping whose power is determined by the bit count. %Emerging \ac{adc} technologies, particularly neuromorphic \acp{adc} using memristors, can greatly vary in power and performance for the same bit count. 
This discrepancy between hardware and algorithmic approaches motivates the design of acquisition schemes that are both task-aware and hardware-conscious, aiming to combine recent algorithmic developments with the capabilities brought forward by memristor technology for neuromorphic \acp{adc}.

% \textcolor{red}{Need to add a paragraph about noise and deep learning for noisy memristors}

In this work, we propose a learning-aided framework for task-based memristive \acp{adc}, which simultaneously considers the signal processing task alongside the power characteristics and inherent controllability of memristors. Our approach integrates memristor-based  \acp{adc} into a power-efficient, task-based hardware-conscious acquisition system. Unlike conventional hardware designs that view \ac{adc} operation as a uniform mapping, our design leverages the adaptive nature of memristive \acp{adc} to optimize performance based on a downstream task, while accounting for the unique controllability and inherent stochasticity of the memristors.

We first present a model that captures the dynamics of the memristor-based \ac{sar} \acp{adc}, focusing on their quantization mapping and noisy configurability while maintaining uniform sampling. We convert this system into a machine learning model, allowing us to optimize the entire acquisition chain, that includes parametric analog pre-processing and digital classification. 
We propose a specialized learning method that accounts for both power constraints and task-specific requirements. Our approach integrates: $(i)$ differentiable approximations of the memristive \ac{adc} mapping; $(ii)$ a dedicated loss function to prevent quantization region collapse due to the indirect relationship between memristor conductance and the resulting \acp{adc}; and $(iii)$ a two-stage learning process that incorporates noise injection and power-aware knowledge distillation to address the inherent stochasticity of memristors. % noise injection during training and knowledge distillation techniques to handle the inherent noise induced by analog circuitry, incorporating strategies from noisy neural networks and specialized regularizers.
Our experimental study encompasses a diverse set of signals, including synthetic signals, classification from microwave imaging~\cite{del2020learned}, and classification based on RF communication signals~\cite{RFclassification}. The results of our numerical trials consistently demonstrate that our combined hardware and algorithmic design significantly enhances the power-accuracy trade-off compared to traditional uniform \acp{adc}. %This approach connects the fields of machine learning and neuromorphic engineering, leveraging the advantages of both to create a more efficient and effective acquisition system.

The rest of this article is organized as follows: Section~\ref{sec:System Model} describes the neuromorphic \ac{adc} and outlines the task-based acquisition systems. Section~\ref{sec:Learning} proposes our data-driven deep task-based system design, while Section~\ref{sec:Experimental Study} experimentally evaluates the system. Finally, Section~\ref{ssec:conclusions} concludes the paper.

%----------------------------------------------------------------------------------------
%	System Model
%----------------------------------------------------------------------------------------
%\vspace{-0.2cm}
\section{System Model}
\label{sec:System Model}
%\vspace{-0.15cm}
In this section we present the system model of neuromorphic task-based \acp{adc}. We begin by describing the \ac{adc} components in Subsection~\ref{ssec:ADCs}, after which we discuss the overall acquisition system in Subsection~\ref{ssec:overallsystem}, and formulate the design problem in Subsection~\ref{ssec:Problem}.

%%%%%%%%%%%%%%%%%%%%%%%%%%%
%%%	Neuromorphic ADCs
%%%%%%%%%%%%%%%%%%%%%%%%%%%
\subsection{Neuromorphic ADC}
\label{ssec:ADCs}
%\vspace{-0.1cm}
\subsubsection{Circuitry}
\begin{figure}
    \centering
    \includegraphics[width=\linewidth]{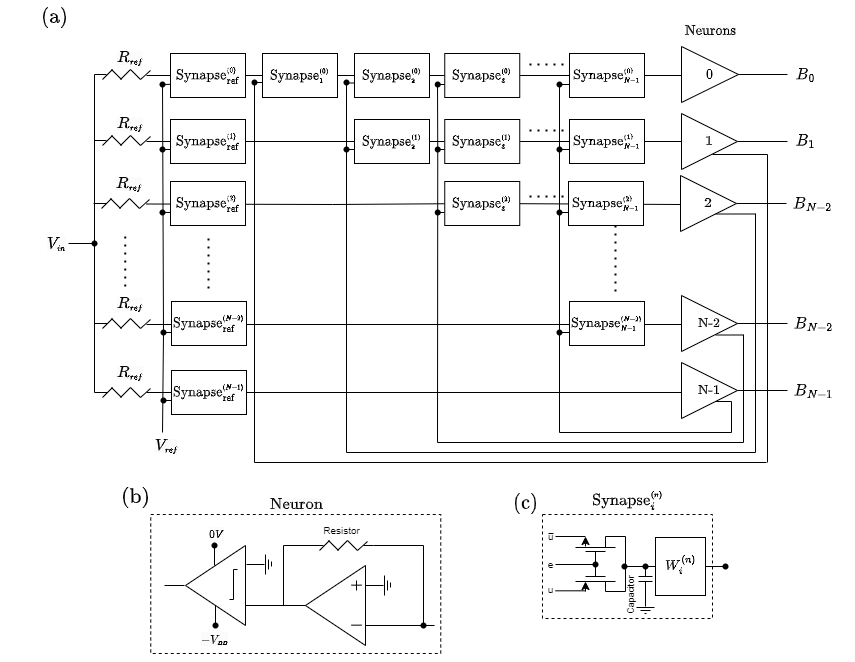}
    \caption{Neuromorphic \ac{sar} \ac{adc} schematic: $(a)$ Overall circuit; $(b)$ Neuron module; $(c)$ Memristive synapse.}
    \label{fig:Neuromorphic ADC Circuit}
\end{figure}

The key component in the signal acquisition chain is the \ac{adc}. We focus on the recently proposed family of {\em neuromorphic \ac{sar} \acp{adc}}. \ac{sar} ADC is a popular architecture with relatively small form factor and power consumption~\cite{razavi1995principles}. Its neuromorphic version implements \ac{sar} \acp{dac} with memristors~\cite{danial2018breaking}, enabling the \ac{adc} to be trainable and not restricted to a fixed mapping. 
Particularly, we focus on neuromorphic implementations of loop-unrolled \ac{sar} \ac{adc}s~\cite{unrolledSAR}, which spatially distribute the decision-making process across multiple comparators and DACs, as depicted in Fig.~\ref{fig:Neuromorphic ADC Circuit}.

% \textcolor{red}{Additionally, our neuromorphic \ac{sar} \ac{adc} resembles the architecture of loop-unrolled \ac{sar} \ac{adc}s~\cite{unrolledSAR}, which spatially distributes the decision-making process across multiple comparators and DACs, as opposed to the traditional temporal approach using a single DAC and register.}

% Paragraph - SAR ADC mapping for a fixed memristor
The principle behind the \ac{sar} ADC is the method of successive approximation, which operates similarly to a binary search algorithm. The four core components of the SAR ADC are the comparator, \ac{sh} circuit, \ac{dac}, and the successive approximation register.
The operation of a SAR ADC is divided into two main phases: the sampling phase and the conversion phase. During the sampling phase, the \acl{sh} circuit acquires and holds the analog input signal value. Once the sampling phase is complete, the conversion phase begins, during which the SAR ADC employs a binary search algorithm to determine the digital output.

%The conversion process starts with the \acl{msb} set to 1 and all other bits set to 0. The DAC uses this digital value to produce an analog voltage that corresponds to the binary representation. This generated analog voltage is then compared with the input signal voltage using the comparator. If the DAC output is less than the input voltage, the MSB remains set to 1; otherwise, it is reset to 0. This operation is repeated in an iterative manner for each bit from the \acl{msb} (denoted $B_{N-1}$) to the \acl{lsb} (denoted $B_0$), refining the approximation at each step.
In each successive step, the SAR sets the next significant bit to 1, adds the previously determined bits, and then the DAC generates a new analog voltage. This new value is again compared with the input signal, and the corresponding bit is either retained as 1 or reset to 0 based on the comparison result. This process continues in an iterative manner for each bit from the \acl{msb} (denoted $B_{N-1}$) to the \acl{lsb} (denoted $B_0$), refining the approximation at each step.
One can formulate the bit construction of analog input voltage $V_{\rm {in}}$ as 
\begin{equation}
\label{eqn:SAR1}
    B_n = \text{sign}\left(V_{\rm {in}} - V_{\text{ref}}^{(n)}\big(\{ B_i\}_{i=n+1}^{N-1}\big)\right), %\quad n=0,\ldots, N-1,
\end{equation}
where the sign function represents the comparator,
%essentially the action of the comparator, that produces '1' if the input voltage is higher than the weighted sum and '0' otherwise, 
and $V_{\text{ref}}^{(n)}$ is obtained from $\{B_{i}\}\in{\{\pm1\}}$ via the \ac{sar} circuitry.

%The neuromorphic architecture depicted in Fig.~\ref{fig:Neuromorphic ADC Circuit}c, uses a single voltage-controlled memristor~\cite{VoltageMemristor} that accepts three voltage inputs: $u$, $\bar{u} = -u$, and the enable signal $e$. These inputs govern the synaptic weight by dictating the selection between input $u$ or $\bar{u}$, thereby altering the writing voltage that is applied through the source terminal of the transistors, and modulating the synaptic state between the high and low resistance states. The enable signal $e$ controls the gates of both MOSFET transistors in the synapse circuit and can take on three values: zero, $V_{\rm DD}$ (positive voltage), or $-V_{\rm DD}$ (negative voltage). $V_{\rm DD}$ is defined as the supply voltage, which is the maximum voltage the ADC can convert. When $e$ is zero, both transistors are non-conducting, effectively disconnecting the synapse. When $e$ is $V_{\rm DD}$, only the NMOS transistor is conducting, applying the writing voltage $-V_w$ through the NMOS transistor, which decrease the memristance. On the contrary, when $e$ is $-V_{\rm DD}$, only the PMOS transistor is conducting, applying the writing voltage $V_w$ (positive voltage) through the PMOS transistor, which increases the memristance. This dynamic adjustment of memristance is essential for the synapse's ability to learn and adapt in the neural network ADC architecture.

The memristive neuromorphic architecture realizes the \ac{sar} circuitry using configurable memristors with trainable (unitless) parameters $\myMat{W}\triangleq \big\{W_{\rm ref}^{(n)}, \{W_{i}^{(n)}\}_{i=n+1}^{N-1}\big\}_{n=0}^{N-1}$ (that are based on its tunable resistors ratios~\cite{VoltageMemristor}). These memristors, which are incorporated into the \ac{adc} via the memristive synapse circuits (Fig.~\ref{fig:Neuromorphic ADC Circuit}c), are dynamically adjustable, and affect the bit conversion in \eqref{eqn:SAR1} by controlling the reference voltage such that
\begin{equation}
\label{eqn:refVolt}
     V_{\text{ref}}^{(n)}\big(\{ B_i\}\big) =\left( W_{\rm ref}^{(n)}  + \sum_{i=n+1}^{N-1}W_{i}^{(n)}\cdot \frac{B_i + 1}{2}\right)V_{\rm w}.
\end{equation}
In  \eqref{eqn:refVolt}, $V_{\rm w}$ is the reference voltage resolution, defined as the supply voltage (the maximum voltage the ADC can convert) divided by $2^{N}$. The reference voltage, in combination with the ADC resolution, determines the minimum voltage difference that the ADC can discern.
Note that the memristor setting only affects the quantization aspect of the \ac{adc}, while sampling is assumed to be uniform and above Nyquist. 

% Paragraph - power consmption
%
\subsubsection{Power}
The power consumption of memristive \acp{adc} is comprised of three terms~\cite{danial2018breaking}:

\begin{enumerate}[label={P\arabic*}]
    \item \label{itm:IntegPower} {\bf Neural Integration Power:}
The first term is the power lost on the feedback resistor of each neuron during the decision-making process. It is given by
\begin{subequations}
    \label{eqn:IntPower}
\begin{equation}
\label{eqn:PowerInt}
P_{\text{int}}^{(n)} = \left(V_{\text{in}}-W_{\text{ref}}^{(n)}V_{\text{ref}}-\sum_{i=n+1}^{N-1}W_{i}^{(n)}{V_i}\right)^2/R_{\text{ref}},
\end{equation}
where $V_n$ represent the $n$th bit in voltage, i.e., $V_n = V_{\rm ref}$ if $B_n=1$ and $V_n = 0$ if $B_n=-1$. Here, $R_\text{ref}$ is the feedback resistor.
%This equation solves the ADC quantization for each neuron, as described in the neural network configuration. 
The total neural integration power dissipated across all neurons is the sum of the individual integration powers:
\begin{equation}
\label{eqn:SumPowerInt}
P_{\text{int}} = \sum_{i=0}^{N-1}P^{(i)}_{\text{int}}.
\end{equation}
\end{subequations}
\item \label{itm:SynapsePower} {\bf Synapse Power:} 
The second term encapsulates the power consumed at the synapses for every neuron within the system. It is given by
\begin{subequations}
    \label{eqn:SynPower}
\begin{equation}
\label{eqn:PowerSyn}
P_{\text{syn}}^{(n)} = \!\left({{V^2_{\text{in}}}}+{W_{\text{ref}}^{(n)}{V^2_{\text{ref}}}}+\!\sum_{i=n+1}^{N-1}{{W_i^{(n)}}}V^2_i\right)/{R_{\text{ref}}}.
\end{equation}
The total synapse power is the sum of the power consumed by each bit, namely, 
\begin{equation}
\label{eqn:SumSynPower}
P_{\text{syn}} = \sum_{i=0}^{N-1}P_{\text{syn}}^{(i)}.
\end{equation}
\end{subequations}
\item \label{itm:ActPower}
{\bf Activation Power:}
The last power term is the activation power. The activation power corresponds to the energy dissipated through the comparators and \acl{opa} at the system's sampling frequency.
This power source is constant and negligible relatively to the resistor based components\cite{danial2018breaking}.

% Specifically, when using a $0.18 \mu m$ CMOS process at the transistor's transition frequency, the activation power is $P^{(n)}_{\text{act}} = 3 \mu {\rm W}$ \cite{danial2018breaking}.

\end{enumerate}

 The dominant sources of power consumption are typically \ref{itm:IntegPower} and \ref{itm:SynapsePower} \cite{danial2018breaking}. Moreover, while \ref{itm:ActPower} is constant for a given hardware and sampling frequency, \ref{itm:IntegPower} and \ref{itm:SynapsePower} are affected by the memristor's configuration, i.e., the parameters $\myVec{W}$.  

% Paragraph - noise odelling
%
\subsubsection{Memristor Noise}
Memristive \acp{adc} enable dynamically adjustable conversion mappings through controllable parameters $\myVec{W}$, which simultaneously influence the digital representation via \eqref{eqn:SAR1}-\eqref{eqn:refVolt} and the power consumption via \ref{itm:IntegPower}-\ref{itm:SynapsePower}. Despite the potential for achieving flexible and power-efficient \acp{adc}, current memristor technology implementations are typically characterized by non-negligible noise, leading to uncertainty and deviations in the memristor settings \cite{devicevairabiltiy}.

While most resistor technologies inherently exhibit some level of noise, such as temperature sensitivity, the dynamic variability of conventional floating gate-based Flash memristors introduces unique forms of inconsistency, known as {\em write} and {\em read} noises~\cite{LoaiNoise}. {\em Write noise} arises from the stochastic nature of the electron injection process during programming, resulting in a statistical spread of resistance states. Conversely, {\em read noise} stems from conductance instability during read operations, causing fluctuations in the weights that vary each time the value is read.

A common and effective model for the variability in the memristor weights, represents these fluctuations as Gaussian noise \cite{Joshi2020, ErrorEst, Dupraz}. The noise level is dynamically adjusted based on the standard deviation of the hardware characteristics \cite{Joshi2020}, capturing the combined effects of read and write noise. 
In the context of the considered memristive \ac{adc}, the weights configured, denoted $\widetilde{\myVec{W}}$, differ from the weights utilized by the memristive \ac{adc}, denoted ${\myVec{W}}$, whose entries are given by
\begin{equation}
    \label{noise_modelling}
{W}_{i}^{(n)} = \widetilde{W}_{i}^{(n)} + \epsilon_{i}^{(n)}. %\quad \text{where} \quad N_{i}^{(n)} \sim \mathcal{N}(0, \sigma_{i}^{2^{(n)}}),
\end{equation}
In \eqref{noise_modelling}, $\{\epsilon_i^{(n)}\}$ are mutually independent zero-mean Gaussian noises, where we use  $\sigma_{i}^{{(n)}}$ to denote the standard deviation of $\epsilon_i^{(n)}$. While in principle, this value can vary with the conductance of the memristor, we follow the common modelling employed in \cite{ErrorEst}, and model the variance of the memristive noise as independent of $\tilde{W}_i^{(n)}$
%
%In this formulation, each memristor will have a different noise component with its own variance, \(\sigma_{i}^{2^{(n)}}\). 
The noise terms are also temporally independent, such that on each \ac{adc} operation, the noise terms take a different realization~\cite{Joshi2020}. %This model, illustrated in Fig.~\ref{fig:noisy_weights}, reflects the real-world memristive devices' intrinsic stochasticity and variability. %Utilizing this method, we can simulate analyze how noise affects the reliability and effectiveness of our neuromorphic systems.

\subsection{Task-Based Memristive Signal Acquisition}
\label{ssec:overallsystem} 
%\vspace{-0.1cm}

\begin{figure*}[t]
    \centering
    \includegraphics[width=0.8\textwidth]{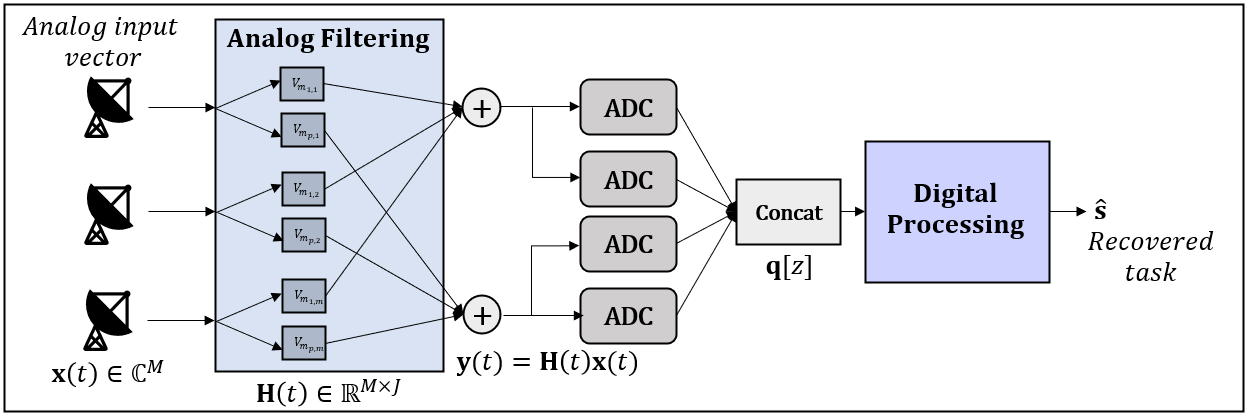}
    \caption{An illustration of a generic task-based acquisition system}
    \label{fig:System Model}
\end{figure*}

The \ac{adc} detailed above can be used as a neuromorphic computing system implementing task-based acquisition, i.e., extracting digital information %(e.g., classification or parameter estimation) 
from analog signals~\cite{shlezinger2018hardware}. The resulting system combines analog pre-processing, signal acquisition, \ac{adc}, and digital processing. The signal acquisition chain is parameterized as artificial neurons that map a multivariate analog signal into a decision, as illustrated in Fig.~\ref{fig:System Model}. 

\subsubsection{Task} 
To model the acquisition system, we consider an input signal acquired at $M$ sensors, denoted  $\myVec{x}(t) \in \mathbb{C}^M$, observed over a time window $t\in[0,T_{\max})$.
The system consists of analog filtering, memristive SAR ADC, and digital processing. %Our goal, is to jointly learn and optimize these components from data. 
The system shown in Fig.~\ref{fig:System Model} is versatile and can be applied to various scenarios such as regression, classification, and detection. In our work, we conduct several experiments with different setups, focusing primarily on a finite classification task. Specifically, we aim to predict the unknown task vector $s\in\mathcal{S}$, which takes values in a finite set ,i.e., $|\mySet{S}|< \infty$. Prediction is based on the input vector $\{\myVec{x}(t)\}_{t\in[0,T_{\max})}$, which is related via the conditional distribution $\mathcal{P}_{x|s}$. This setting encapsulates various scenarios such as, e.g., \ac{mimo} detection in wireless communications~\cite{shlezinger2018asymptotic}, \ac{mimo} radar~\cite{xi2021bilimo}, and ultrasound beam-forming in medical imaging~\cite{huijben2020learning}, where efficient and accurate signal acquisition and detection is critical.

\subsubsection{Analog Processing} The analog signal measured at the sensor array at time instance $t$, denoted $\myVec{x}(t)$, is mapped using the analog circuitry through a memory-less, linear mapping parameterized by the vector $\myVec{\theta}_{\rm 1}$. This mapping transforms $\myVec{x}(t)$ into $J$ continuous-time signals, denoted as $\{y_j(t)\}_{j=1}^J$.  These signals represent the output after the analog filtering procedure. Our study focuses on linear analog filtering, where the relationship between the input and output of the analog processing can be described as
\begin{equation}
\label{eqn:AnalogMap}
\myVec{y}(t) = \myMat{H}(\myVec{\theta}_{1}) \myVec{x}(t).
\end{equation}
In \eqref{eqn:AnalogMap}, $\myMat{H}(\myVec{\theta}_{1}) \in \mathbb{R}^{M \times J}$ is the linear analog mapping, and $\myVec{\theta}_{\rm 1}$ are the parameters of the analog network.

\subsubsection{Analog-to-Digital Conversion} 
In the next step, $\myVec{y}(t)$ is converted into a digital representation via the trainable memristive ADCs. Let $T_s$ be the ADC sampling interval. The output of the $j$th ADC is the discrete time $N$-bit sequence $q_j[z] \in \{0, \ldots, 2^{N}-1\}$, where $z \in \{0, \ldots, Z \}$ with $Z \triangleq \lfloor T_{\max}/T_s \rfloor$. Here, $q_j[z]$ is the digital representation of $y_j(z \cdot T_s)$, and $B_{j,n}[z]$ is its $n$th bit, which is computed bit-by-bit via \eqref{eqn:SAR1} with $V_{\rm in} = y_j(z \cdot T_s)$. Accordingly, the $n$th bit used for representing the $j$th entry of the $z$th multivariate sample is obtained as
\begin{equation}
B_{j,n}[z] = {\rm sign}\left(y_j(z \cdot T_s) - \sum_{i=n+1}^{N-1} W^{(n)}_i B_{j,i}[z]\right),
\label{eq:Qsign}
\end{equation}
where each bit is a function of the previous $n+1, \ldots, N$ bits. Then, all the bits are summed together to form the final digital output of the $j$th \ac{adc} at time $z$, via
\begin{equation}
    q_j[z] = \sum_{i=0}^{N-1} \frac{B_{j,i}[z] + 1}{2} \cdot 2^i.
\end{equation}
As each $J$-dimensional multivariate sample (out of $Z$ samples) is represented using $N$ bits, it holds that the overall number of bits used in acquisition is  $Z \cdot J \cdot N$.

\subsubsection{Digital Processing} The sequence of digital representations denoted $\myVec{q}[0], \ldots, \myVec{q}[Z]$ is  processed in digital to recover $\myVec{s}$. To model the overall acquisition chain as a machine learning model, we focus on digital processing using a \ac{dnn} with trainable parameters $\myVec{\theta}_2$. Depending on the specific application requirements, this network can vary in design. In our examples, we use both fully connected linear networks and \ac{cnn}. However, as we focus on classification, the \ac{dnn} has a softmax output layer, such that its output is a $|\mySet{S}|\times 1$ probability vector denoted
\begin{equation}
    \myVec{p}_{\myVec{\theta}}\left(\{\myVec{x}(t)\}\right) =  f_{\myVec{\theta}_{\rm 2}}(\myVec{q}[0], \ldots, \myVec{q}[Z]),
    \label{eqn:SoftEstimate}
\end{equation}
where the resulting estimate is denoted by:
\begin{equation}
\label{eqn:Estimate}
    \hat{\myVec{s}} = \myVec{s}_{k^\star} \in \mySet{S} \leftrightarrow k^\star = \mathop{\arg\max}\limits_{k\in \{1,\ldots,|\mySet{S}|\}}\left[ \myVec{p}_{\myVec{\theta}}\left(\{\myVec{x}(t)\}\right) \right]_k.
\end{equation}
In \eqref{eqn:SoftEstimate}-\eqref{eqn:Estimate}, $\myVec{\theta}_{\rm 2}$ is the trainable parameters of the digital DNN, while $\myVec{\theta} \triangleq \{\myVec{\theta}_1, \widetilde{\myMat{W}}, \myVec{\theta}_2\}$ is the overall system parameters. %The output of the last layer of the digital network is a conditional probability distribution $\mathcal{P}_{\hat{s}|x}$, representing the probability of each possible label given the input.

%%%%%%%%%%%%%%%%%%%%%%%%%%%
%%%	Problem Formulation
%%%%%%%%%%%%%%%%%%%%%%%%%%%
%%\vspace{-0.1cm}
\subsection{Problem Formulation}
\label{ssec:Problem}
%\vspace{-0.1cm}
Our goal is to design a task-based acquisition system with memristive \acp{adc} for recovering $\myVec{s}$. We aim to optimize the system's accuracy while meeting an overall power constraint $P_{\max}$. This involves setting the system parameters $\myVec{\theta} = \{\myVec{\theta}_1, \widetilde{\myMat{W}}, \myVec{\theta}_2\}$ based on the following objective
\begin{subequations}
    \label{eqn:problem}
    %\vspace{-0.1cm}
\begin{align}
    \label{eqn:problemobj}
&\min \left( \mathbb{P}( \hat{\myVec{s}}(\myVec{\theta)} \neq \myVec{s}) \right) \\
&\text{subject to} \quad  \mathbb{E} \left\{ P_{\text{syn}}(\myMat{W}) + P_{\text{int}}(\myMat{W}) \right\}\leq P_{\text{max}},
%\vspace{-0.1cm}
    \label{eqn:problemPower} 
\end{align}
\end{subequations}
where the expectation in the constraint is taken with respect to the distribution of $\myVec{W}$ (given $\widetilde{\myMat{W}}$), whose stochasticity stems from the memristor noise. 
In \eqref{eqn:problem}, we explicitly state the dependency of the estimate and power terms on the system parameters. The proposed framework unifies the \ac{adc} configuration along with both the analog and digital processing in light of the overall objective of maximizing accuracy under constrained power budget.
%The entire system is trained as a joint entity, composed of several segments, each with unique weights that can be fine-tuned. These weights, defined as 
The controllable system parameters, $\myVec{\theta}$, adjust and calibrate the individual components of the system, ultimately shaping the overall performance.

% Specifically, the weights of the analog network, $\myVec{\theta}_1$, correspond to the adjustable parameters of the analog pre-processing, e.g., tunable phase shifters~\cite{lavi2023learn}. The second category of weights pertains to the \ac{adc}. These correpsond to the configurable memristance, represented by $W^{(n)}_{i}$, which allows to manage the state and behavior of the synapse units.
%Lastly, we introduce the digital network weights $\myVec{\theta}_2$, which serve as the key parameters of the digital neural network. 
The task-oriented acquisition system is designed to learn how to infer $s$ from $\myVec{x}(t)$ using a training set denoted as:
\begin{equation}
\label{eqn:Dataset}
\mathcal{D}=\{\{\myVec{x}^{(r)}(t)\}_{t \in [0, T_{\max})}, \myVec{s}^{(r)}\}_{r=1}^{R},
\end{equation}
which comprises $R$ instances of inputs along with their corresponding task vectors. The learning process during training includes acquiring the parameters of the analog filter $\myMat{H}(\boldsymbol{\theta}_1)$, the weights $W^{(n)}_{i}$ of the SAR ADC, and the weights $\myVec{\theta}_2$ of the digital neural network.
The data is used to tune $\myVec{\theta}$ based on \eqref{eqn:problem} as proposed in the following section.

{\bf Challenges:}
In designing neuromorphic task-based \acp{adc}, one encounters three primary challenges to be addressed:
\begin{enumerate}[label=C\arabic*]
    \item \textbf{The non-differentiable nature of the ADCs:} \label{challenge:C1} The ADCs are integral components of task-based acquisition. To train the system using data-driven methods, it is necessary to perform gradient descent from the output layer back to the input layer. However, the presence of non-differentiable ADCs in the middle of the system poses a significant challenge, as it restricts the applicability of conventional deep learning techniques.
    
    \item \textbf{The relationship between quantization decision regions and memristor weights:} \label{challenge:C2} In our practical system, the memristor weights directly influence and dictate the decision regions. This is in contrast to the synthetic formulations considered in previous works, such as \cite{shlezinger2022deep}, where the ADCs do not directly parameterize each decision region separately.
    
    \item \textbf{The stochasticity induced by the memristive noise:} \label{challenge:C3} Memristors are inherently noisy components, and this noise induces stochasticity in the system. This stochastic behavior complicates the reliable functioning and accurate modeling of the system.
\end{enumerate}
%
%To tackle these challenges, we  design the overall acquisition system using machine learning techniques, as detailed in the following section.

%----------------------------------------------------------------------------------------
%	Learning Neuromorphic ADCs
%----------------------------------------------------------------------------------------
%\vspace{-0.2cm}
\section{Learning Task-Based Memristive ADCs}
\label{sec:Learning}
%\vspace{-0.15cm}

\begin{figure*}[t]
    \centering
    \includegraphics[width=0.8\textwidth]{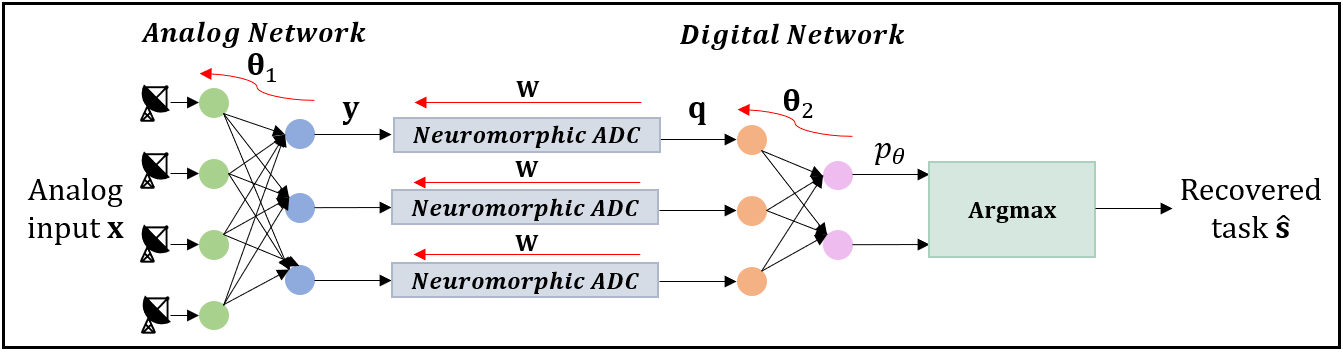}
    \caption{Machine learning model of the neuromorphic task-based acquisition system.}
    \label{fig:Machine learning model}
\end{figure*}

The formulation of \eqref{eqn:problem} motivates tuning the system parameters by treating it as a machine learning model. Consequently, the system depicted in Fig.~\ref{fig:System Model} can be represented by the trainable discriminative machine learning model~\cite{shlezinger2022discriminative} shown in Fig.~\ref{fig:Machine learning model}. The first layer of this model corresponds to the analog processing, parameterized by $\myVec{\theta}_1$; the second layer consists of the memristive \acp{adc} with parameters $\myMat{W}$; and the subsequent layers encompass the digital processing, parameterized by $\myVec{\theta}_2$. To optimize $\myVec{\theta}$ using the dataset $\mathcal{D}$, we formulate a loss function designed based on \eqref{eqn:problem} while accounting for \ref{challenge:C2} in Subsection~\ref{ssec:loss}; develop the corresponding learning algorithm that tackles \ref{challenge:C1} and \ref{challenge:C3} in Subsection~\ref{ssec:Training}; and provide a discussion in Subsection~\ref{ssec:Discussion}.

\subsection{Loss Function}
\label{ssec:loss}
Problem  \eqref{eqn:problem} includes two performance measures: accuracy (via the objective \eqref{eqn:problemobj}) and power (via the constraint \eqref{eqn:problemPower}). Accordingly, we formulate a loss that accounts for both measures. Moreover, to account for the challenging in relating the decision regions to the memristor weights in simple form (\ref{challenge:C2}), we add a dedicated regularization to the \ac{adc} parameters. 

The loss achieved by a signal acquisition chain with parameters $\myVec{\theta}$ evaluated on data set $\mySet{D}$ is given by 
\begin{equation}
\mathcal{L}_{\mySet{D}}(\myVec{\theta}) =  \mathbb{E}\left\{\mathcal{L}_{\mySet{D}}^{\text{CE}}(\myVec{\theta}) + \alpha \cdot  \mathcal{L}^{\text{reg}}({\myMat{W}})+\beta \cdot P_{\mySet{D}}(\myMat{W})\right\}.
\label{eq:loss}
\end{equation}
In \eqref{eq:loss}, $\mathcal{L}_{\mySet{D}}^{\text{CE}}$ is 
the averaged cross-entropy loss (as we focus on classification tasks), given by 
\begin{equation}
\!\mathcal{L}_{\mySet{D}}^{\text{CE}}(\myVec{\theta}) \!=\! -\sum_{r=1}^{\mathcal{D}} \sum_{k=1}^{|\mathcal{S}|} 1_{s^{(r)}=s_k} \log\left( \left[ \myVec{p}_{\myVec{\theta}}\left(\{\myVec{x}^{(r)}(t)\}\right) \right]_k \right),
\label{eqn:cross_entropy}
\end{equation}
with $\myVec{p}_{\myVec{\theta}}(\cdot)$ being the probability vector produced by the digital \ac{dnn} defined in \eqref{eqn:SoftEstimate}, while $1_{(\cdot)}$ denotes the indicator function. The term $ P_{\mySet{D}}$ is the averaged total power (computed via \eqref{eqn:problemPower}) when applied to $\mySet{D}$, and $\alpha$ and $\beta$ are hyperparameters, where the latter is empirically tuned to guarantee that \eqref{eqn:problemPower} holds. The stochastic expectation in \eqref{eq:loss} is taken with respect to the conditional distribution of $\myMat{W}$ given the setting of $\widetilde{\myMat{W}}$ in $\myVec{\theta}$ (which is the distribution of $\myVec{\epsilon}$).

% \begin{figure*}[t]
%     \centering
%     \begin{subfigure}{\textwidth}
%         \centering
%         \includegraphics[width=\textwidth]{figs/reg_graph.png}
%         \caption{Decision regions illustration}
%         \label{fig:weight_reg}
%     \end{subfigure}
    
%     \vspace{0.4cm}
    
%     \begin{subfigure}{.4\textwidth}
%         \centering
%         \includegraphics[width=\linewidth]{figs/8_levels.png}
%         \caption{8 Distinct quantization levels - no merging issue}
%         \label{fig:8_levels}
%     \end{subfigure}
%     \hfill
%     \begin{subfigure}{.4\textwidth}
%         \centering
%         \includegraphics[width=\linewidth]{figs/6_levels.png}
%         \caption{6 Distinct quantization levels - merging issue exists}
%         \label{fig:6_levels}
%     \end{subfigure}
    
%     \vspace{0.5cm}
    
%     \caption{Illustrations of decision regions and quantization levels of a 3-bit memristive \ac{adc}: $(a)$ decision regions and their correspondence with memristor weights; $(b)$ a setting that maintains the full range of the ADC; and $(c)$ a setting with collapsing decision regions, reducing the effective quantization levels. \textcolor{red}{we need to reorder this figure}}
%     \label{fig:quantization_levels}
% \end{figure*}

\begin{figure*}[t]
    \centering
    \includegraphics[width=\textwidth]{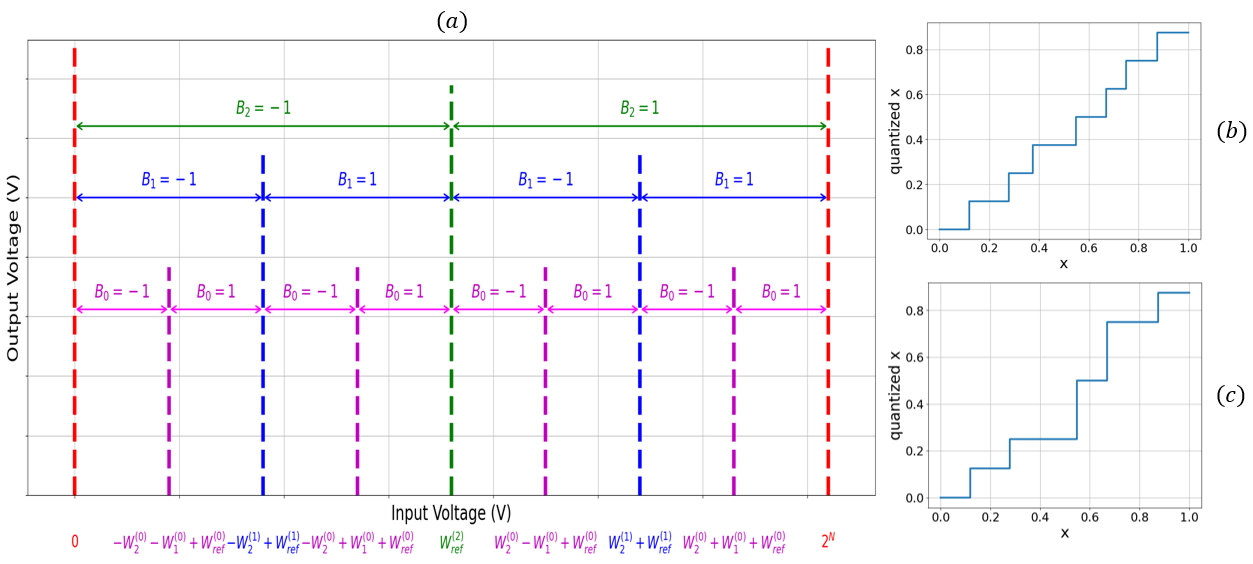}
    \caption{Illustrations of decision regions and quantization levels of a 3-bit memristive \ac{adc}: $(a)$ decision regions and their correspondence with memristor weights; $(b)$ a setting that maintains the full range of the ADC; and $(c)$ a setting with collapsing decision regions, reducing the effective quantization levels.}
    \label{fig:quantization_levels}
\end{figure*}

 The regularization term $\mathcal{L}^{\text{reg}}$ in \eqref{eq:loss} is designed to tackle \ref{challenge:C2} and prevent resolution collapse, i.e., preserve $2^N$ quantization levels. To illustrate the need for this regularizer, we visualize in Fig.~\ref{fig:quantization_levels}(a) the relationship between each edges of the decision region of the \ac{adc} and the memristor weights. Different unique combinations of weights dictate the position of each edge, affecting the output range for each input range. When decision regions overlap, the full resolution of the \acp{adc} is not exploited, leading to resolution collapse in which some digital representations merge, undermining overall performance. This is illustrated in Fig.~\ref{fig:quantization_levels}(c), where the absence of regularization causes the weights to collapse, resulting in only 6 distinct values instead of the expected 8 for a 3-bit case. In contrast, Fig.~\ref{fig:quantization_levels}(b) shows a mapping learned with the regularization term, which effectively prevents this issue, exploiting the full range of the \ac{adc}.

To formulate the regularizer, let ${b}_i^{(j)} \in \pm 1$ be the $i$th bit of the digital encoding of the $j$th decision region, with $i \in \{0, \ldots, N-1\}$ and $j \in \{1, \ldots, 2^N\}$. Let $l_j \in [0, N-1]$ denote the level of the $j$th decision region, i.e., the index $i$ of the least significant bit for which ${b}_i^{(j)} = -1$ and ${b}_i^{(j+1)} = 1$. For instance, for a 3-bit \ac{adc}, as shown in Fig.~\ref{fig:quantization_levels}(a), the level of the first decision region is $l_1 = 0$, since ${b}_0^{(1)} = -1$ while ${b}_0^{(2)} = 1$. 

Using these notations, the right edge of the $j$th decision region, denoted by $E_j(\myMat{W})$, is given by:
\begin{equation}
E_j(\myMat{W}) = W_{\text{ref}}^{(l_j)} + \sum_{i=l_j + 1}^{N-1} b_i^{(j)}W_i^{(l_j)}.
\end{equation}
Accordingly, we set the regularization term to be:
\begin{equation}
 \mathcal{L}^{\text{reg}}(\myMat{W}) = \sum_{j=1}^{2^N-1} e^{\max\{E_{j}(\myMat{W}) - E_{j+1}(\myMat{W}), 0\}},
\end{equation}
thus penalizing overlaps between decision regions, thereby enhancing the robustness and accuracy of the \ac{adc}. This regularization encourages the left edge of each decision region to remain lower than the right edge, resulting in distinct ranges.
 
% TODO CONTINUE FROM HERE
%%%%%%%%%%%%%%%%%%%%%%%%%%%
%%%	Training
%%%%%%%%%%%%%%%%%%%%%%%%%%%
%%\vspace{-0.1cm}
\subsection{Training}
\label{ssec:Training}
Based on the loss formulated in \eqref{eq:loss}, the training algorithm seeks to tune the parameters $\myVec{\theta}$ based on the following empirical loss optimization problem
\begin{equation}
\label{eqn:TrainingOpt}
\myVec{\theta}^*=\mathop{\arg \min}\limits_{\myVec{\theta} =\{\myVec{\theta_2},\widetilde{\myVec{W}},\myVec{\theta_1}\}}\mathcal{L}_{\mySet{D}}(\myVec{\theta}).
\end{equation}

The formulation in \eqref{eqn:TrainingOpt} motivates tuning $\myVec{\theta}$ using training algorithms based on, e.g., mini-batch \ac{sgd}. However, doing so requires handling the two main distinctions between the machine learning representation of the task-based acquisition system and conventional machine learning architectures: the non-differentiable nature of the \ac{adc} operation \ref{challenge:C1}, and the stochasticity induced by the memristor noise \ref{challenge:C3}. Therefore, to formulate our training method, we first introduce differentiable approximations of the \ac{adc} mapping to tackle  \ref{challenge:C1}, after which we propose a knowledge distillation based learning method inspired by noisy neural network training techniques to cope with \ref{challenge:C3}. 

\subsubsection{Differentiable ADC Approximation}
\begin{figure}
    \centering
    \includegraphics[width=0.8\linewidth]{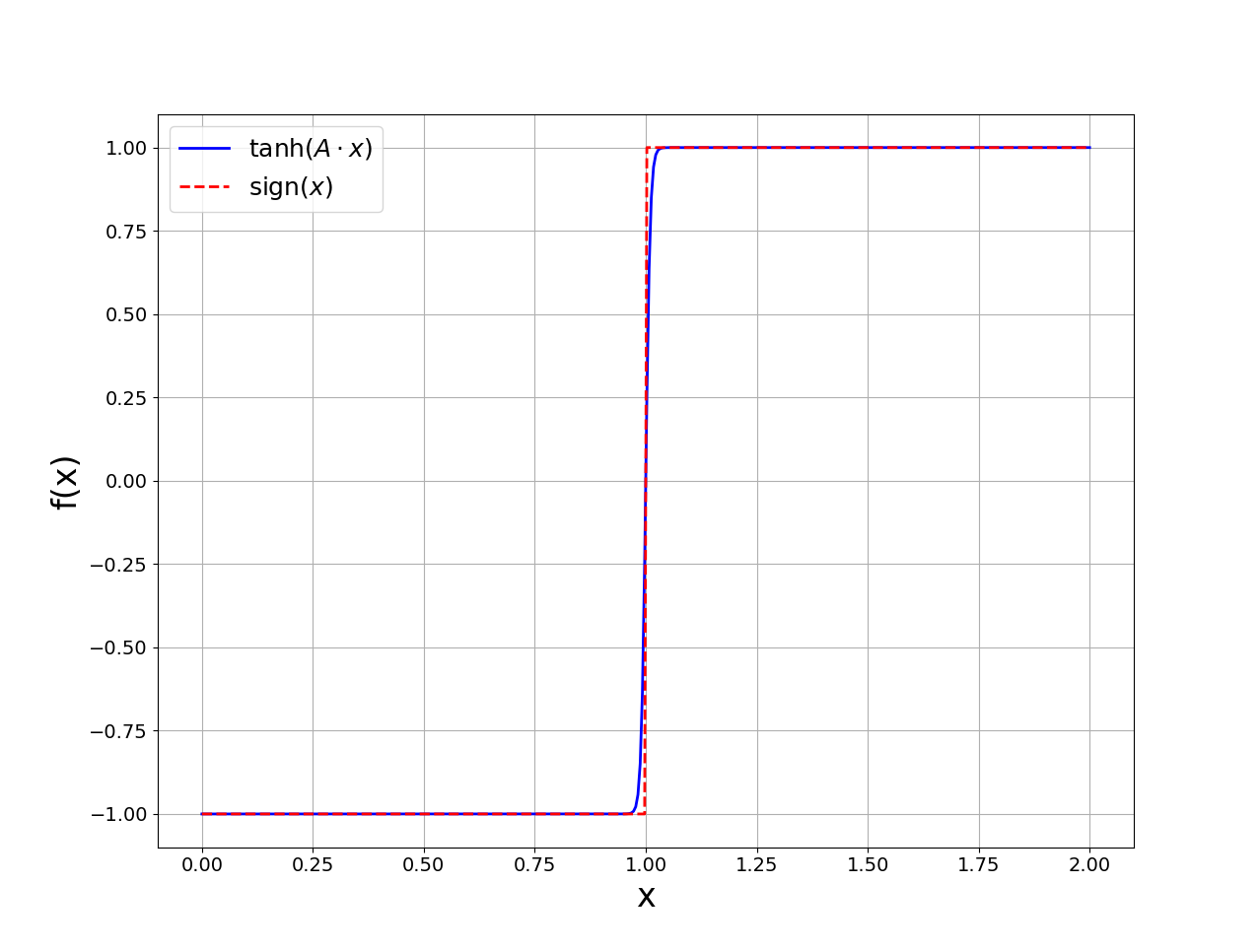}
    \caption{Approximation of $\text{sign}(x)$ using $\tanh(Ax)$.}
    \label{fig:sign_appro}
\end{figure}

To address the non-differentiable nature of the ADCs (\ref{challenge:C1}), we follow the approach of\cite{agustsson2017soft, shlezinger2021deep, shlezinger2022deep} and use a soft approximation to compute the gradients. Specifically, when computing the loss gradients, we approximate the sign function in \eqref{eqn:SAR1} with 
\begin{equation}
 B_{j,n}[z] = \tanh\left(A\left(y_{j}(z T_{\rm s}) - V_{\rm ref}^{(n)}(\{B_{j,i}[z]\}_{i>n})\right)\right).
\label{eqn:approxSign}
\end{equation}
As illustrated in Fig.~\ref{fig:sign_appro}, the approximation \eqref{eqn:approxSign} closely matches the non-differentiable \ac{adc} rule in \eqref{eqn:SAR1}.

By adjusting the hyperparameter $A$, the term $\tanh(A \cdot x)$ can be made to closely approximate the $\text{sign}(x)$ function, while still allowing gradients to propagate effectively. Increasing the value of $A$ makes the $\tanh$ function more closely resemble the $\text{sign}$ function, thereby making \eqref{eqn:approxSign} more representative of the ADC behavior. However, a larger $A$ also reduces the gradient magnitudes, which may hinder the learning process by causing the gradients to approach zero. 
%This presents a trade-off: while a larger $A$ enhances the realism of the system, it may also impede effective optimization. Therefore, $A$ must be carefully tuned based on the specific requirements of the system and the dataset.

\subsubsection{Training as a Noisy \texorpdfstring{\ac{dnn}}{DNN}}
Next, we tackle the stochasticity induced during inference by noisy memristors (\ref{challenge:C3}). To that aim, we treat the architecture as a form of a {\em noisy \ac{dnn}}~\cite{disti,BayesianApp,HardwareAware,NoisyRNN}. Inspired by recent advances in training of noisy \acp{dnn}, we incorporate two techniques for enabling reliable training or such machine learning models: noisy training and knowledge distillation with a noise-free teacher.

{\bf Noisy Training}: 
To enable the learning procedure to operate reliably with noisy memristors, we incorporate noise during the training phase, using {\em noise injection}  \cite{HardwareAware,NoisyRNN}. This method integrates noise into the DNN training process to help the network maintain consistent performance across a wide range of noise levels during inference. Specifically, we inject Gaussian noise into the memristive weights during training to make them more resilient to the noise encountered during inference. %By training under noisy conditions, the resulting memristor conductance weights are better prepared for the stochastic nature of the hardware.

\begin{figure*}
    \centering
    \includegraphics[width=0.8\textwidth]{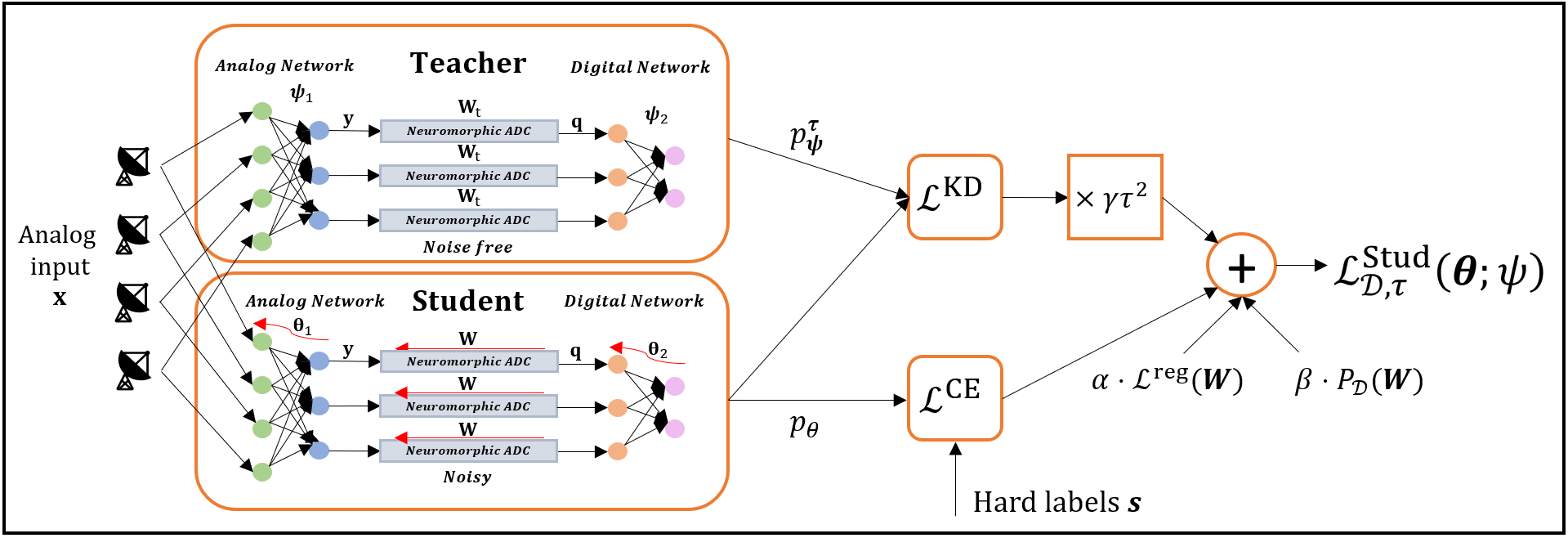}
    \caption{Teacher-student distillation illustration of the overall training loss.  
    }
    \label{fig:Distillation model}
\end{figure*}
 
{\bf Knowledge Distillation}: 
Following \cite{disti}, we enhance the training process through knowledge distillation, guiding training from a noise-free teacher model to train a noisy student model \cite{disti}. Knowledge distillation leverages soft labels generated by the teacher model to train the student \ac{dnn}. 
We use the teacher model to guide the task-based acquisition system in carrying out its classification task, and thus we set it to be a trainable task-based acquisition system that does not have to handle with the core constraints imposed on the memristive system -- the need to cope with noise (\ref{challenge:C3}) and the constraint on the power consumption~\eqref{eqn:problemPower}. {Accordingly, the teacher model is a software realization of the {\em noiseless} task-based memristive \ac{adc} with {\em unconstrained power}.}
Specifically, by letting $\myVec{\psi}$ denote the trainable parameters of the teacher, these are trained from the data  $\mySet{D}$ using the following loss function
\begin{equation}
\mathcal{L}_{\mySet{D}}^{\rm Teac}(\myVec{\psi} = \{\myVec{\psi}_2, \myVec{W}_t, \myVec{\psi}_1\}) = \mathcal{L}_{\mySet{D}}^{\text{CE}}(\myVec{\theta}) + \alpha \cdot \mathcal{L}^{\text{reg}}({\myVec{W}_t}). 
\label{eq:basic_loss}
\end{equation}

%In our setup, the teacher model is a noiseless memristive ADC, while the student model is a memristive ADC with noise injected during training. 

Once trained, the noise-free teacher model guides the soft outputs produced by the student, i.e., the noisy and power-constrained task-based \ac{adc}. Following the established practice in knowledge distillation~\cite{gou2021knowledge}, we add a temperature value $\tau$ to control the softness of the probability vectors produced by the teacher. 
To formulate this, let $\myVec{z}_{\myVec{\psi}}\big(\{\myVec{x}(t)\}\big)$ be the logits of the teacher model (i.e., the input to the softmax output layer) with parameters $\myVec{\psi}$ when applied to signal  $\{\myVec{x}(t)\}$. The $\tau$-softened teacher output is given by the vector $\myVec{p}_\myVec{\psi}^{\tau}(\{\myVec{x}(t)\}\big)$, whose $k$th entry (for each  $k\in\{1,\ldots,|\mySet{S}|\}$) is 
\begin{equation}
\left[\myVec{p}_\myVec{\psi}^{\tau}(\{\myVec{x}(t)\}\big)\right]_k =  \frac{\exp \left(\left[\myVec{z}_{\myVec{\psi}}\big(\{\myVec{x}(t)\}\big)\right]_k^\tau/ \tau)\right)}{\sum_j \exp \left(\left[\myVec{z}_{\myVec{\psi}}\big(\{\myVec{x}(t)\}\big)\right]_j^\tau/ \tau)\right)}.
\label{eqn:Temp}
\end{equation}
In \eqref{eqn:Temp}, increasing $\tau$ results in softer labels, providing more nuanced guidance to the student network. By adjusting the temperature, we can generate probability vectors that range from hard labels (low $\tau$) to soft labels (high $\tau$).

The $\tau$-softened teacher output regularizes the training of the student, using an additional cross-entropy term added to the loss. This term encourages the probability estimates of the student to approach those of the teacher, and is given by 
\begin{align}
\mathcal{L}_{\mySet{D}, \tau}^{\text{KD}}(\myVec{\theta}; \myVec{\psi})  = \sum_{r=1}^{\mathcal{D}} \sum_{k=1}^{|\mathcal{S}|} &\left[\myVec{p}_\myVec{\psi}^{\tau}(\{\myVec{x}(t)\}\big)\right]_k \notag \\ 
&\times\log\left( \frac{\left[\myVec{p}_\myVec{\psi}^{\tau}(\{\myVec{x}^{(r)}(t)\}\big)\right]_k}{\left[ \myVec{p}_{\myVec{\theta}}\left(\{\myVec{x}^{(r)}(t)\}\right) \right]_k} \right),
\label{eqn:Distcross_entropy}
\end{align}
The resulting loss used to train the student for temperature parameter $\tau$ using data $\mySet{D}$ and a trained teacher $\myVec{\psi}$ is
\begin{align}
\mathcal{L}_{\mySet{D}, \tau}^{\rm Stud}(\myVec{\theta}; \myVec{\psi}) =\mathbb{E}\Big\{& \mathcal{L}_{\mySet{D}}^{\text{CE}}(\myVec{\theta}) + \gamma \tau^2 \cdot \mathcal{L}_{\mySet{D}, \tau}^{\text{KD}}(\myVec{\theta}; \myVec{\psi}) \notag \\
& + \alpha \cdot \mathcal{L}^{\text{reg}}(\myMat{W})
 + \beta \cdot P_{\mySet{D}}(\myMat{W}) \Big\},
\label{eq:full_loss}
\end{align}
where $\gamma > 0$ is an additional hyperparameters, and the expectation is taken with respect to the memristor noise distribution. 

\subsubsection{Training Algorithm Summary}
The resulting training algorithm thus consists of two stages: The first trains a noise-free teacher model, and the second uses the teacher to train the noisy power-constrained task-based memristive \ac{adc}. The training of the teacher with parameters $\myVec{\psi}$ is based on the loss \eqref{eq:basic_loss}, and is carried out from the data $\mySet{D}$ in \eqref{eqn:Dataset} using conventional gradient-based learning, while computing the gradients with the soft approximations of the \ac{adc} mapping in \eqref{fig:sign_appro}.  The resulting first stage using mini-batch \ac{sgd} is summarized as Algorithm~\ref{alg:algo1}.

\begin{algorithm}
    \caption{Training Teacher Model}
    \label{alg:algo1}
    \SetKwInOut{Initialization}{Init}
    \Initialization{Set hyperparameters $A, \alpha$ and initial  $\myVec{\psi}$; \\ Fix learning rate $\mu>0$ and epochs $i_{\max}$; 
    }
    \SetKwInOut{Input}{Input}  
    \Input{Training set $\mathcal{D}$}  
    \For{$i = 0, 1, \ldots, i_{\max}-1$}{
        Randomly divide $\mathcal{D}$ into $Q$ batches $\{\mathcal{D}_q\}_{q=1}^Q$\;
        \For{$q = 1, \ldots, Q$}{
            Compute batch loss $\mathcal{L}^{\rm Teac}_{\mathcal{D}_q}(\myVec{\psi})$ via \eqref{eq:basic_loss}\;
            Compute $\nabla_{\myVec{\psi}}\mathcal{L}^{\rm Teac}_{\mathcal{D}_q}(\myVec{\psi})$ replacing \eqref{eqn:SAR1} with \eqref{eqn:approxSign}\;
            Update $\myVec{\psi} \leftarrow \myVec{\psi} - \mu\nabla_{\myVec{\psi}}\mathcal{L}^{\rm Teac}_{\mathcal{D}_q}(\myVec{\psi})$\;
        }
    }
    \KwRet{$\boldsymbol{\psi}$}
\end{algorithm}

Algorithm~\ref{alg:algo1} is employed to train the teacher model. Once the teacher model is trained, it serves as a reference to guide the training of the student model, which is both noisy and power-aware. While a common practice in knowledge distillation is to have the teacher more highly parameterized compared to the student~\cite{zhang2021compacting}, here one can also use similar architectures for both (as we do in Section~\ref{sec:Experimental Study}), as the teacher here has an inherent advantage over the student stemming from the absence of noise and power constraints.  

The student training is guided by the loss function in \eqref{eq:full_loss}, where noisy training is employed. Specifically, the stochastic expectation over the memristive noise distribution in \eqref{eq:full_loss} is computed using Monte Carlo sampling, by adding a new i.i.d. realization of $\myVec{\epsilon}$ to each input. An example of this training process, based on mini-batch \ac{sgd}, is detailed in Algorithm~\ref{alg:algo2}. While Algorithm~\ref{alg:algo2} is formulated with fixed hyperparameters, such as the soft-to-hard coefficient $A$ and temperature $\tau$, the method can be extended to allow these values to change during training, as in simulated annealing.

\begin{algorithm}
    \caption{Noisy Power-Aware Training}
    \label{alg:algo2}
    \SetKwInOut{Initialization}{Init}
    \Initialization{Set hyperparameters $A, \alpha, \beta, \gamma, \tau$ and initial  $\myVec{\theta}$; \\ Fix learning rate $\mu>0$ and epochs $i_{\max}$; }
    \SetKwInOut{Input}{Input}  
    \Input{Training set $\mathcal{D}$, trained teacher $\myVec{\psi}$}  
    \For{$i = 0, 1, \ldots, i_{\max}-1$}{
        Randomly divide $\mathcal{D}$ into $Q$ batches $\{\mathcal{D}_q\}_{q=1}^Q$\;
        \For{$q = 1, \ldots, Q$}{
            Apply teacher model $\myVec{\psi}$ to $\mathcal{D}_q$\;
            Sample  $|\mathcal{D}_q|$ i.i.d. realizations of $\myVec{\epsilon}$\;
            Compute batch loss $\mathcal{L}_{\mySet{D}_q, \tau}^{\rm Stud}(\myVec{\theta}; \myVec{\psi})$ via \eqref{eq:full_loss}, evaluating $\mathbb{E}\{\cdot \}$ with Monte Carlo averaging\;
            Compute $\nabla_{\myVec{\theta}}\mathcal{L}_{\mySet{D}_q, \tau}^{\rm Stud}(\myVec{\theta}; \myVec{\psi})$ replacing \eqref{eqn:SAR1} by \eqref{eqn:approxSign}\;
            Update $\myVec{\theta} \leftarrow \myVec{\theta} - \mu\nabla_{\myVec{\theta}}\mathcal{L}_{\mySet{D}, \tau}^{\rm Stud}(\myVec{\theta}; \myVec{\psi})$\;
        }
    }
    \KwRet{$\myVec{\theta}$}
\end{algorithm}

\subsection{Discussion}
\label{ssec:Discussion}
Our proposed system designs task-based \acp{adc} based on the emerging concrete technology of memristive \acp{adc}. We develop a dedicated algorithm to jointly account for the system task and the hardware specificities and the unique power profile of such circuitry. Unlike previous works on task-based acquisition, e.g.,~\cite{shlezinger2018hardware,shlezinger2022deep}, here we directly relate the \ac{adc} configuration into power, faithfully capturing its dependence on the mapping and not just the number of bits. 

Our learning algorithm copes with the core challenges \ref{challenge:C1}-\ref{challenge:C3} we identified as limiting machine learning based designs of such systems. We achieved this by combining 
$(i)$ hardware-oriented differentiable approximations of the \ac{sar} \ac{adc} mapping to handle \ref{challenge:C1};
$(ii)$ a dedicated regularizer to overcome possible resolution collapse stemming from \ref{challenge:C2}; and
$(iii)$ a two-stage learning procedure combining noisy learning with knowledge distillation from an unconstrained noise-free teacher to overcome \ref{challenge:C3}. Combining these hardware-oriented learning techniques allows to learn reliable low-power hardware-compliant task-based \ac{adc} configurations, as we systematically show in Section~\ref{sec:Experimental Study} for various different signals, encompassing  synthetic as well as realistic analog signals. 

The learning procedure via Algorithms~\ref{alg:algo1}-\ref{alg:algo2} can be computationally intensive (representing two \ac{dnn} training operations). Still, they are done offline, i.e., during design stage. Once the the parameters $\myVec{\theta}$ are learned, the resulting system realizes the task-based memristive acquisition system, as illustrated in Fig.~\ref{fig:System Model}. The learning procedure requires prior knowledge of the memristive noise distribution, which can be obtained from measurements or from established models, e.g.,~\cite{LoaiNoise,Joshi2020, ErrorEst, Dupraz}.

We focus on memristive \ac{sar} \acp{adc}, being a popular neuromorphic architecture for power-efficient acquisition. As the tunable conductance of memristive \acp{adc} affects the quantization mapping, our focus is on fixed uniform sampling. However, our joint hardware-algorithmic design can also be extended to other forms of  \acp{adc}, based on e.g., flash architectures~\cite{FlashMem}, as well as tunable non-uniform sampling~\cite{neuhaus2020task, shlezinger2022deep}. Nonetheless, such extensions necessitate revisiting the power formulation which is affected by the hardware and sampling interval. Moreover, our design does not account for the power of the analog circuitry, which is generally not fixed and can be optimized~\cite{tasci2022robust}. Furthermore, we focus on settings where the signals are acquired for a specific  task. One can potentially extend such designs to multiple tasks as a form of, e.g., multi-task learning~\cite{zhang2021survey}. We leave these extensions for future work. 

% Additional ADCs

% power of analog mapping

% implement for on device learning

%----------------------------------------------------------------------------------------
%	Experimental Study
%----------------------------------------------------------------------------------------
%\vspace{-0.2cm}
\section{Experimental Study}
\label{sec:Experimental Study}
%\vspace{-0.15cm}
Here, we evaluate task-based signal acquisition, in terms of accuracy-power trade-off.
We evaluate our neuromorphic task-based system trained using Algorithm~\ref{alg:algo2}, coined {\em Memristive distillation}, comparing it to systems with uniform \acp{adc}, termed {\em Uniform}. We also evaluate our power-aware learning by evaluating additional three forms of  neuromorphic task-based systems:
$(ii)$ Training without distillation, i.e., $\gamma = 0$ in \eqref{eq:full_loss}, referred to as {\em Memristive noisy training}; 
$(ii)$ Training without accounting for the stochasticity of the memristors, coined {\em Memristive noisy inference}; and
$(iii)$ Task-based acquisition without noise, termed {\em Memristive noise free}. 

We consider three case studies\footnote{The source code is available at \url{https://github.com/talvol/Deep-task-based-acquisition-with-trainable-ADCs/}}: 
an experimental study with synthetic signal (Subsection~\ref{ssec:SynProb}), which allows us to evaluate our design steps in a controlled setting; 
an imaging setting involving handwritten digit recognition (Subsection~\ref{ssec:MNISTProb}), representing classification-based microwave imaging setups as in \cite{del2020learned}; 
and the classification of communication RF signals representing different waveforms~\cite{RFclassification} (Subsection~\ref{ssec:rf_signal}).
Power calculations  are based on the power analysis of Subsection~\ref{ssec:ADCs} with memristor values implementing either the learned quantization mapping, or a uniform setup (for non-learned \acp{adc}).

\subsection{Case Study: Synthetic Signal Model}
\label{ssec:SynProb}

% \begin{figure*}[t]
%     \centering
%     \begin{subfigure}{0.45\textwidth}
%         \centering
%         \includegraphics[width=\linewidth]{figs/beta.png}
%         \caption{Hyperparameter tuning $\beta$}
%         \label{fig:beta_test}
%     \end{subfigure}
%     \hfill
%     \begin{subfigure}{0.45\textwidth}
%         \centering
%         \includegraphics[width=\linewidth]{figs/noise.png}
%         \caption{Noise level impact}
%         \label{fig:noise_test}
%     \end{subfigure}
%     \vskip\baselineskip
%     \begin{subfigure}{0.45\textwidth}
%         \centering
%         \includegraphics[width=\linewidth]{figs/acc_syn_dis.png}
%         \caption{Accuracy}
%         \label{fig:acc_syn}
%     \end{subfigure}
%     \hfill
%     \begin{subfigure}{0.45\textwidth}
%         \centering
%         \includegraphics[width=\linewidth]{figs/power_syn_dis.png}
%         \caption{Power}
%         \label{fig:power_syn}
%     \end{subfigure}
%     \caption{Synthetic task.}
%     \label{fig:Synthetic_task}
% \end{figure*}

\begin{figure}
    \centering
    \includegraphics[width=0.45\textwidth]{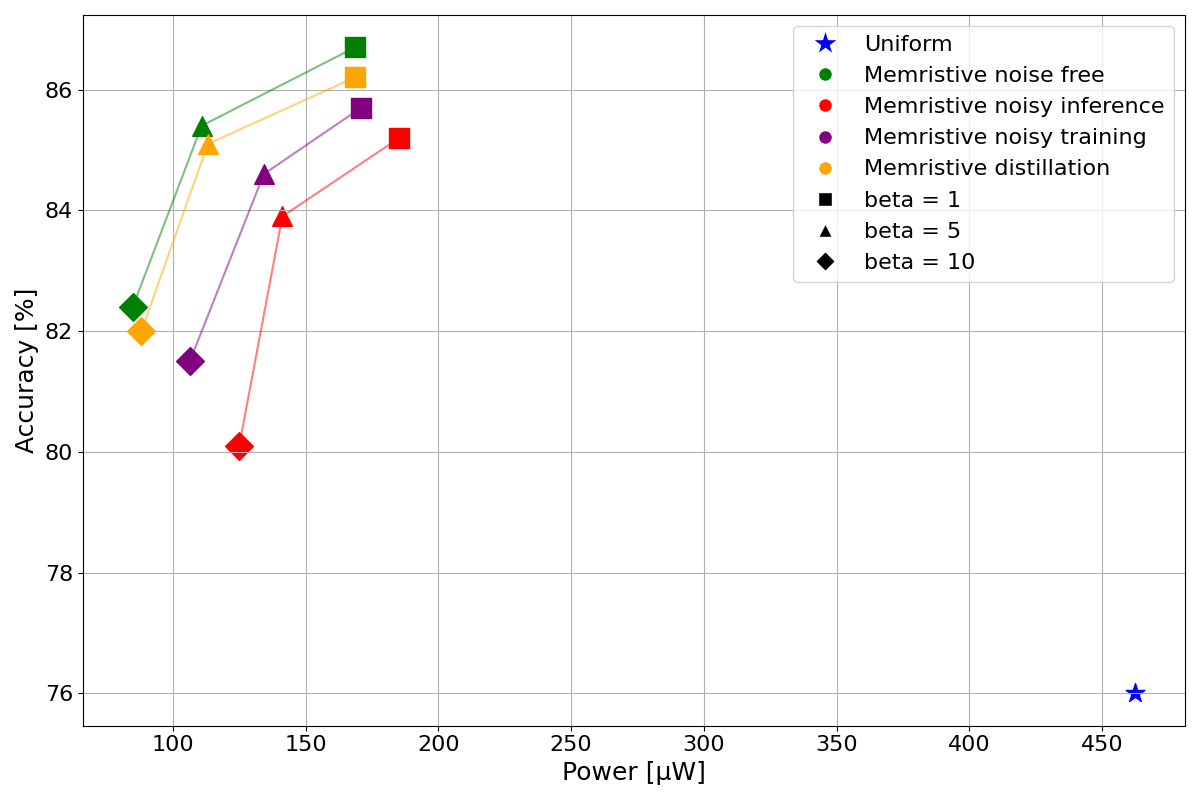}
    \caption{Accuracy-power trade-offs by tuning $\beta$.}
    \label{fig:beta_test}
\end{figure}

\begin{figure}
    \centering
    \begin{subfigure}{.45\textwidth}
        \centering
        \includegraphics[width=\linewidth]{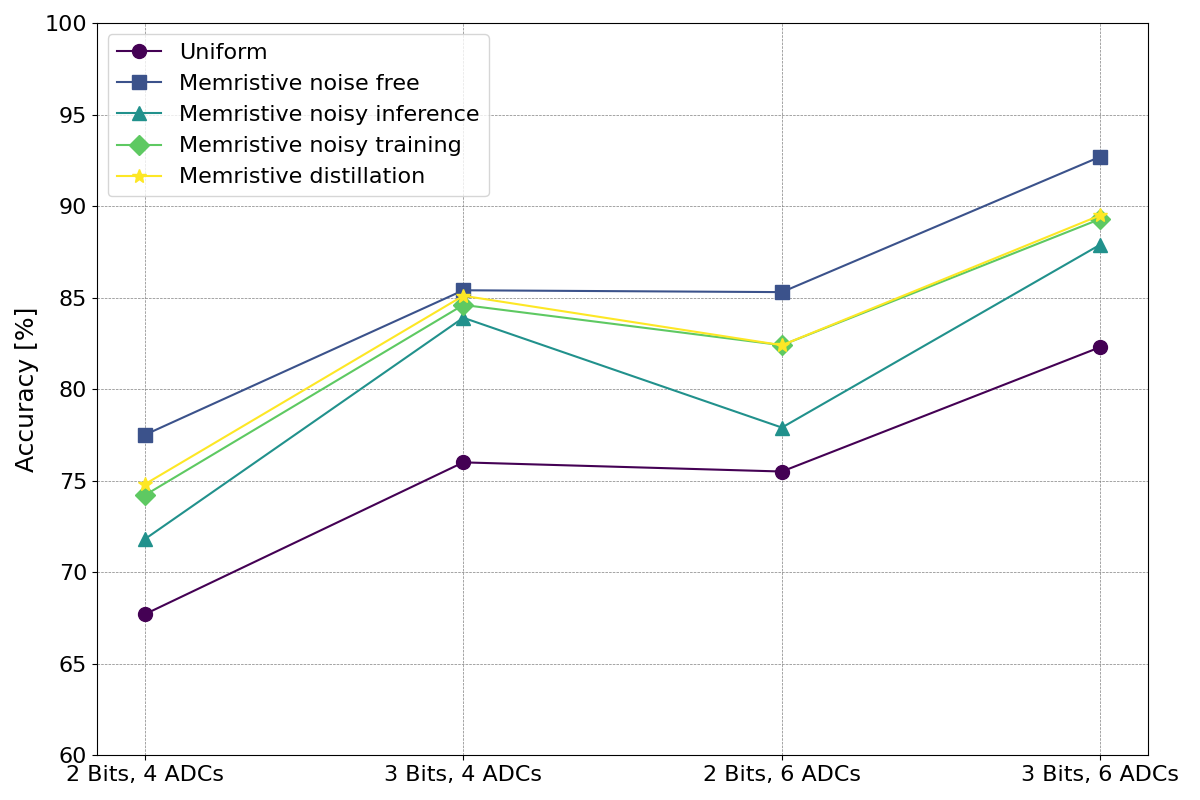}
        \caption{Accuracy}
        \label{fig:acc_syn}
    \end{subfigure}

    \vspace{0.4cm}
    
    \begin{subfigure}{.45\textwidth}
        \centering
        \includegraphics[width=\linewidth]{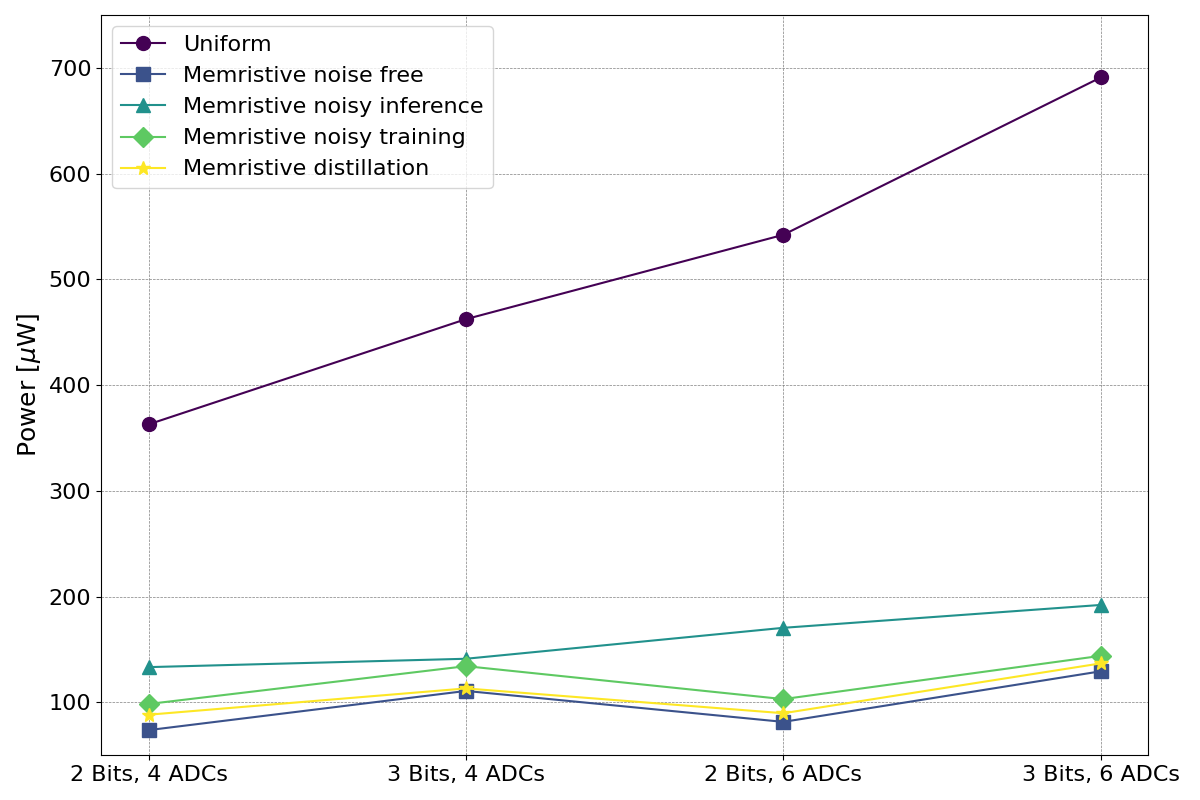}
        \caption{Power}
        \label{fig:power_syn}
    \end{subfigure}
    \vspace{0.4cm}
    \caption{Synthetic task.} 
    \label{fig:comparison_synthetic}
\end{figure}

We commence by evaluating our task-based acquisition system in performing detection tasks from signals generated synthetically. The main goal is to illustrate how the suggested system performs as a single integrated unit and show how it can train holistically. The effects of several hyperparameters are also investigated, revealing  the trade-off between prediction accuracy and power reduction. In doing so, we examine the role of hyperparameter tuning and assess the effectiveness of our power-aware task-based memristive \acp{adc}.

\subsubsection{Task Formulation}
We focus on classification tasks where the target $\myVec{s}$ is drawn from $|\mySet{S}| = 32$ classes. Specifically, we write $\myVec{s}$ as a $5\times 1$ vector whose entries take value in $\{-1,1\}$, such that $\mySet{S} = \{-1, 1\}^5$. 
Following \cite{shlezinger2022deep} the task vector undergoes a noisy linear transformation, after which it is measured with $M=16$ sensors over  $T_{\max} = 3$ milliseconds at sampling rate of $1/T_{\rm s} = 10^3$ Hz. 
Specifically, the received signal $\myVec{x}(t)$ is related to $\myVec{s}$ via 
\begin{equation}
\myVec{x}(t) = \myMat{G}(t) \myVec{s}+\myVec{w}(t).
\end{equation}
Here, $\myVec{w}(t)$ is additive white Gaussian noise with unit variance, while the measurement matrix  $\myMat{G}(t)$ represents spatial exponential decay with temporal variations, and its entries are
\begin{equation}
(\mathbf{G}(t))_{a,b} = \sqrt{\rho}(1+0.5\cos(2\pi t/T_{\rm s})e^{-|a-b|},
\end{equation}
with $\rho > 0$ denoting the \ac{snr}. 
The data set is comprised of $n_t = 2 \times 10^4$ samples.

\subsubsection{Acquisition System}
 In the conducted experiment, the task-based ADC conversion is comprised of an analog linear mapping, a SAR ADC, and a digital DNN. 
 The analog pre-processing $\myMat{H}(\myVec{\theta}_1)$ is a $M\times J$ Fourier matrix, whose parameters are the selected discrete frequencies, i.e., its $(m,j)$th entry is 
 \begin{equation}
     [\myMat{H}(\myVec{\theta}_1)]_{m, j} = \exp\big( -2\pi \sqrt{-1} \frac{m [\myVec{\theta}_1]_j}{M}\big).
 \end{equation}
 The digital DNN consists of an initial layer of dimensions $2J \times 64$, followed by a ReLU activation function, an intermediate layer of dimensions $64 \times 32$, and a softmax output layer.
The output of the implemented network is a probability vector spanning $\mathcal{S}^k$. Training is based on the ADAM optimizer \cite{Adam} with a learning rate of 0.001 and a batch size of 1024.

\begin{table}
    \vspace{2em}
    \centering
    \begin{adjustbox}{width=\columnwidth}
    \begin{tabular}{|l|c c|c c|c c|c c|}
        \hline
        & \multicolumn{2}{|c|}{\makecell{Memristive \\ noisy inference}} 
        & \multicolumn{2}{|c|}{\makecell{Memristive \\ noisy training}} 
        & \multicolumn{2}{|c|}{\makecell{Memristive \\ distillation}} 
        & \multicolumn{2}{|c|}{\makecell{Memristive \\ noise free}} \\
        \hline
         $\sigma_{i}^{(n)}$
        & \makecell{Acc. \\ (\%)} 
        & \makecell{Power \\ ($\mu$W)} 
        & \makecell{Acc. \\ (\%)} 
        & \makecell{Power \\ ($\mu$W)} 
        & \makecell{Acc. \\ (\%)} 
        & \makecell{Power \\ ($\mu$W)} 
        & \makecell{Acc. \\ (\%)} 
        & \makecell{Power \\ ($\mu$W)} \\
        \hline
        0.01 & 82 & 162.37 & 84.7 & 150.23 & {\bf 85.8} & {\bf 144.09} & & \\
          0.05 & 78 & 173.19 & 83.4 & 163.21 & {\bf 83.6} & {\bf 159.28} & 85.8 & 137.64 \\
         0.1 & 74.3 & 174.41 & 80.5 & 167.97 & {\bf 80.9} & {\bf 161.23} & & \\
        \hline
    \end{tabular}
    \end{adjustbox}
    \caption{Performance under different noise conditions.}
    \label{tab:memristive_performance}
\end{table}

\subsubsection{Results}
In our first study, reported in Fig.~\ref{fig:beta_test}, we analyze the effect of the hyperparameter $\beta$ on both accuracy and power consumption. As the value of $\beta$ increases, the emphasis of the loss function shifts towards power reduction rather than classification accuracy. This shift results in a decrease in power consumption alongside a reduction in accuracy. The results demonstrate that tuning the $\beta$ hyperparameter allows for a trade-off between power and accuracy, providing flexibility in optimizing for either criterion based on specific application requirements.
We also observe in Fig.~\ref{fig:beta_test} how each ingredient of our power-aware training mechanism, and particularly noisy training and distillation, contributes to improving the achievable accuracy-power trade-offs. The resulting Pareto frontier achieved when training via Algorithm~\ref{alg:algo2}  approach  noise-free systems, and notably improve upon using uniform \acp{adc}.

Table~\ref{tab:memristive_performance} illustrates the impact of noise on accuracy and power consumption. As the noise level increases, there is a negative effect on both metrics: power consumption rises, and accuracy decreases. Notably, when noise is minimal, employing a combination of distillation training and noise injection during training achieves results that are nearly equivalent to the noise-free scenario. This approach reaches comparable accuracy while maintaining slightly higher power consumption. This finding suggests that our power-aware learning effectively mitigates the adverse effects of memristor noise, preserving high accuracy with modest increases in power consumption.

To evaluate the performance of our memristive ADC network, we compared it to a uniform ADC network with varying numbers of ADC units (4, 6) and different bit resolutions (2, 3). Our evaluation focused on accuracy and power consumption across several scenarios, including the impact of noise. The memristor noise variance is set to $\sigma_{i}^{(n)} = 0.01$ following~\cite{ErrorEst}. The results, illustrated in Figs.~\ref{fig:acc_syn} and ~\ref{fig:power_syn}, demonstrate that our trained system consistently outperforms the uniform system in both accuracy and power efficiency. The neuromorphic setup, which employs clean memristor-based ADCs, shows significant reductions in power consumption and notable increases in accuracy. When noise is present during inference, there is a drop in accuracy and an increase in power consumption. However, accounting for noise during training improves both metrics, as the system becomes better prepared to handle noise during inference. Our distillation approach further enhances performance, achieving the lowest power consumption and highest accuracy. These findings highlight the effectiveness of our trained neuromorphic task-based acquisition system with memristive ADCs, demonstrating its potential for resource-constrained applications.

\subsection{Case Study: Handwritten Digit Recognition}
\label{ssec:MNISTProb}

\begin{figure}
    \centering
    \begin{subfigure}{.45\textwidth}
        \centering
        \includegraphics[width=\linewidth]{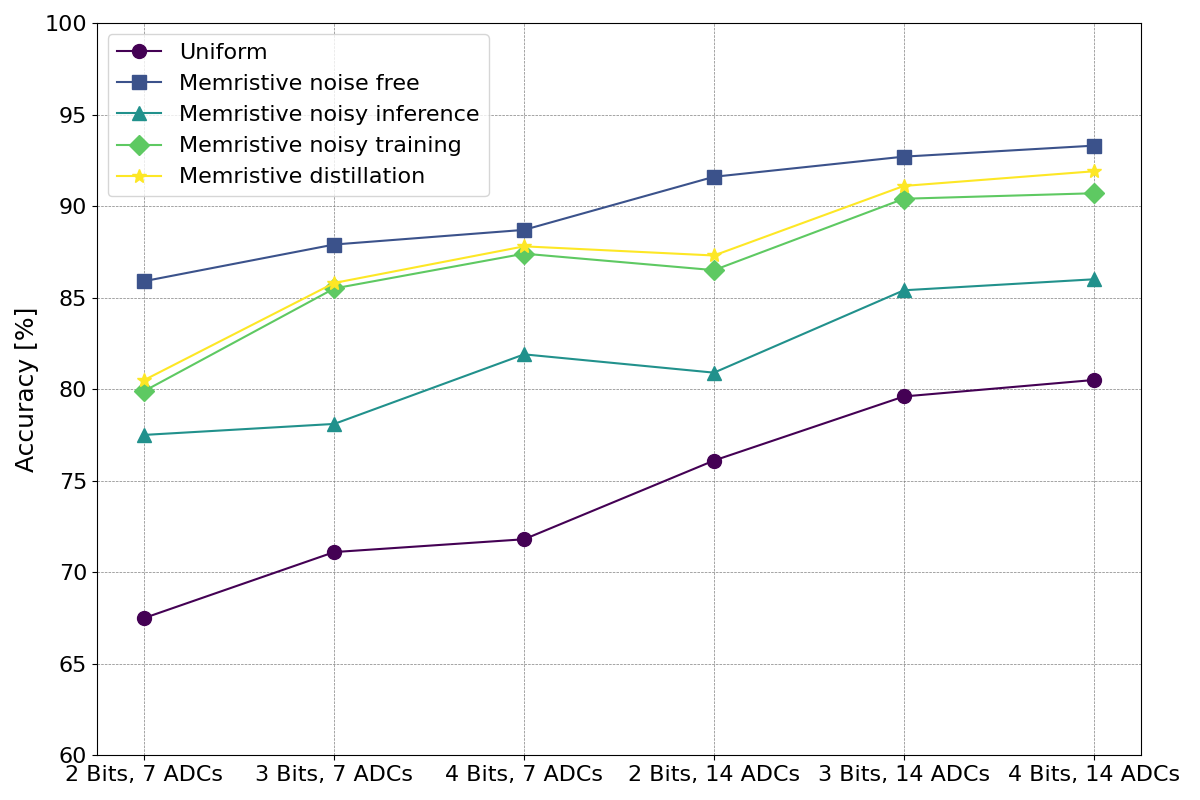}
        \caption{Accuracy}
        \label{fig:accuracy_mnist}
    \end{subfigure}

    \vspace{0.4cm}
    
    \begin{subfigure}{.45\textwidth}
        \centering
        \includegraphics[width=\linewidth]{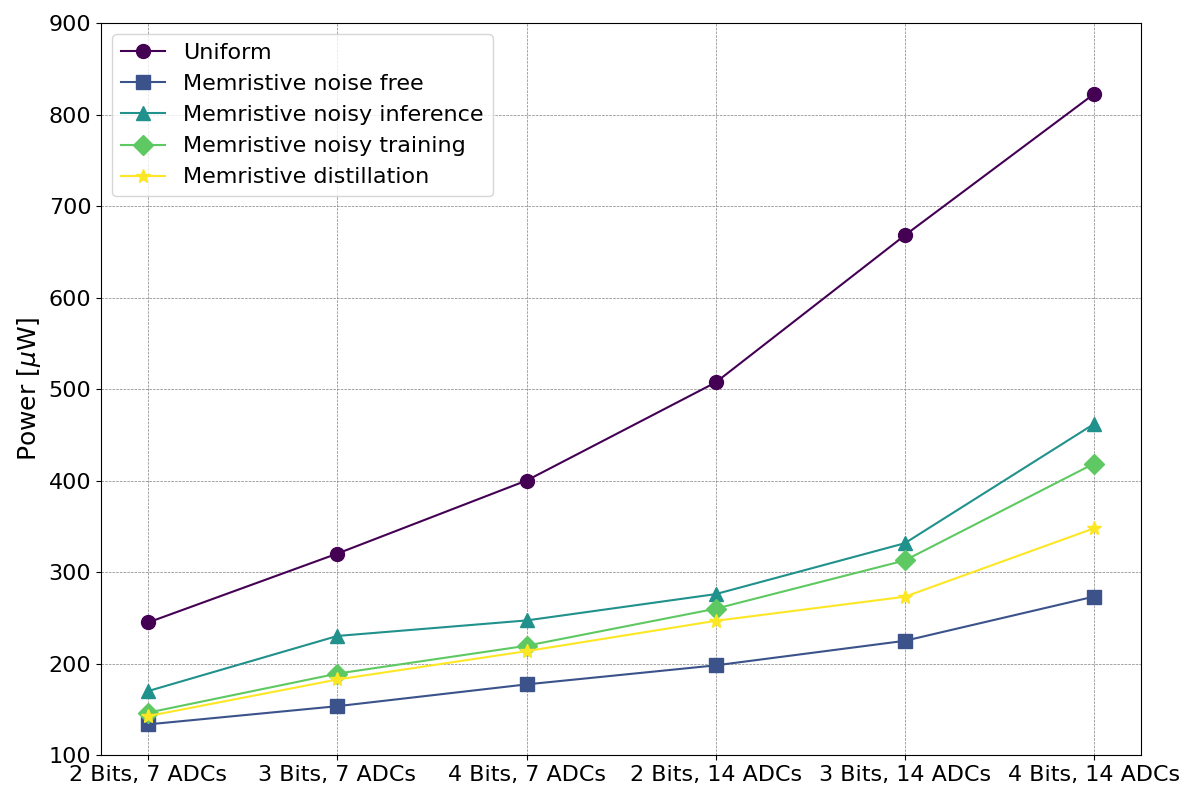}
        \caption{Power}
        \label{fig:power_mnist}
    \end{subfigure}
    \vspace{0.4cm}
    \caption{Handwritten digit recognition task.} 
    \label{fig:comparison_mnist}
\end{figure}

We proceed to evaluate the proposed memristive task-based acquisition system for a scenario emerging from imaging applications. We  focus on handwritten digit recognition, following the setting proposed in \cite{del2020learned} for classification based on microwave imaging.

\subsubsection{Task Formulation}
We consider a single snapshot, where $\myVec{x}(t)$ corresponds to a $28\times 28$ image taken from the MNIST data set. The gray-scale image is initially reshaped into a one-dimensional vector of length $l = 784$. The task is to classify these images into one of 10 possible digit classes. In this experiment, we  evaluate the system's performance under higher noise conditions, setting $\sigma_{i}^{(n)} = 0.3$.

\subsubsection{Acquisition System}
The vectorized image is subsequently processed by an analog network represented by the \ac{dct} matrix $\myMat{H}(\myVec{\theta}_1) \in \mathbb{C}^{M\times J}$, where $J$ is the number of SAR ADCs in the system. Each entry in the matrix is written as
\begin{equation}
[\myMat{H}(\myVec{\theta}_1))]_{m, j} = \phi_{m,j} \cdot \cos\left(\frac{\pi}{M}((j+0.5)m)+[\myVec{\theta}_1]_{m,j}\right),
\end{equation}
where $\phi_{m,j}$ is a normalization factor.  

The output of the analog layer is then processed by an array of SAR ADCs. The converted digital signals then enter the digital layer of the network for classification. The DNN configuration consists of an input layer of dimensions $2J \times 64$, followed by a ReLU activation function, an intermediate layer of dimensions $64 \times 10$, and a softmax output layer. %The class corresponding to the highest probability is then selected as the predicted class of the input vector.

\subsubsection{Results}
In order to evaluate the performance of our memristive ADC network, we perform a comparison between our ADC network to a network using uniform ADC with varying numbers of ADC units (7,14), across several scenarios and differing bit numbers (2,3,4).

The experimental results, illustrated in Figs.~\ref{fig:accuracy_mnist} and ~\ref{fig:power_mnist}, highlight the significant improvements in power efficiency and accuracy achieved by our power-aware learning method. The uniform ADC setup, which does not utilize neuromorphic trainable \ac{adc} mapping, consistently exhibits the highest power consumption and lowest accuracy, underscoring its limitations. In contrast, the neuromorphic setup, which employs clean memristor-based ADCs, shows a substantial reduction in power consumption and a notable increase in accuracy.
In particular, the presence of memristive noise yields a slight increase in power consumption and a drop in accuracy, demonstrating the impact of noise on performance. However, when noise is accounted for during training, the system shows improved power efficiency and accuracy, as it is better prepared to handle noise during inference.

The distillation approach further enhances both power efficiency and accuracy, achieving the lowest power consumption and highest accuracy among all setups. 
Notably, when comparing the configurations of 4 bits with 7 ADCs and 2 bits with 14 ADCs, we observe a unique phenomenon. With 14 ADCs, the system inherently has more noisy components, which, when noise is present, leads to a noticeable drop in accuracy compared to the 4-bit, 7-ADC setup. This indicates that having a higher number of noisy ADCs can outweigh the benefits of increased resolution, highlighting the importance of balancing the number of ADCs and their noise levels.

\subsection{Case Study: RF Signal Classification}
\label{ssec:rf_signal}
We conclude our experimental study with the classification of communications RF signals, encompassing a variety of modulation types. We use the dataset detailed in ~\cite{RFclassification}, which contains signals representing $|\mySet{S}|=18$ different transmission modes commonly found in the HF band. %The task is to accurately classify a received analog RF signal based on its waveform.

\begin{figure*}
    \centering
    
    \begin{subfigure}{.45\textwidth}
        \centering
        \includegraphics[width=6cm, height=4cm]{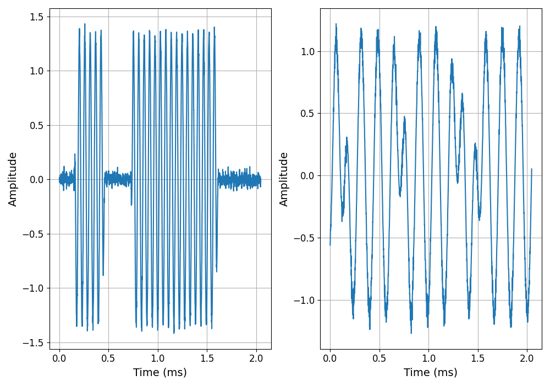}
        \caption{Received RF signals}
        \label{fig:received_rf}
    \end{subfigure}
    \hfill
    \begin{subfigure}{.45\textwidth}
        \centering
        \includegraphics[width=6cm, height=4cm]{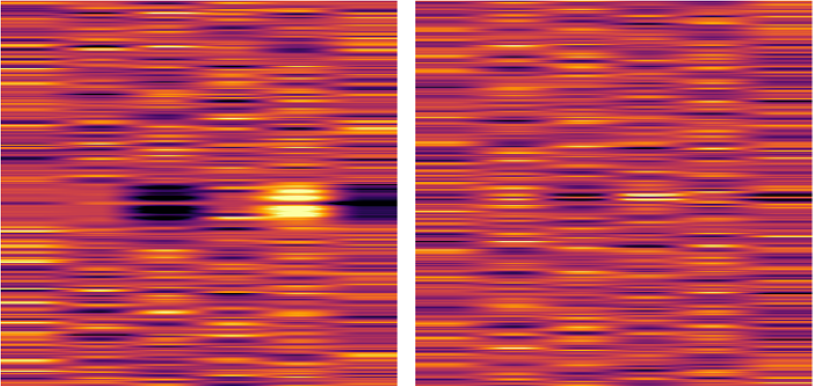}
        \caption{Spectrograms of RF signals}
        \label{fig:spectrogram_rf}
    \end{subfigure}
    \vspace{0.5cm}
    \begin{subfigure}{\textwidth}
        \centering
        \includegraphics[width=0.8\linewidth]{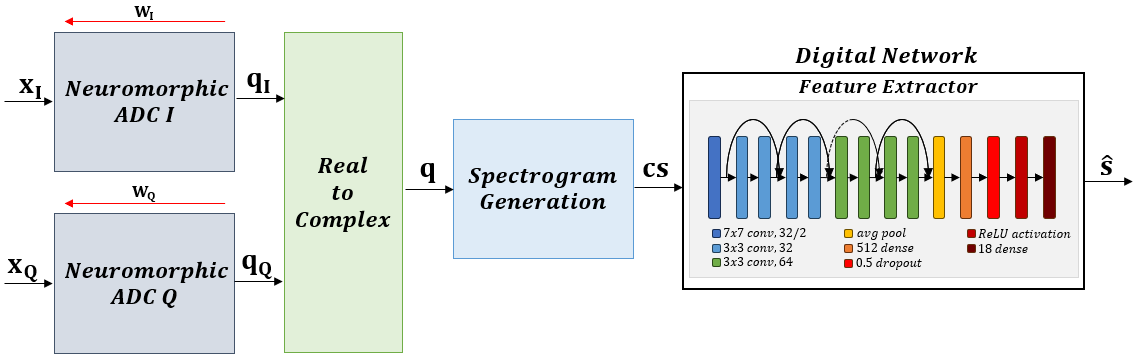}
        \caption{Schematic overview of the RF signal classification system.}
        \label{fig:rf_overview}
    \end{subfigure}
    
    \vspace{0.5cm}
    \caption{Task-based acquisition for RF signal classification, including  representative examples of RF signals in $(a)$ the time domain and $(b)$ spectrogram form; and $(c)$ Schematic system overview.}
    \label{fig:RF}
\end{figure*}

\subsubsection{Task Formulation}
The system task is to classify between $18$ different transmission modes based on their unique modulation characteristics. The RF signal dataset contains $172,800$ signal vectors, with each vector comprising $2048$ complex IQ samples, sampled at a rate of $1/T_{\rm s}=6$ kHz. The signals are synthesized by modulating speech, music, and text, followed by the introduction of impairments such as Gaussian noise, Watterson fading, and random frequency and phase offsets.

\subsubsection{Acquisition System}
The acquisition system proposed in this study is based on the architecture proposed  for LoRa signal classification in \cite{LoRa}, with modifications to adapt it to the RF signal dataset. The received complex time-domain signal $\myVec{x}(t)$, visuailzed in Fig.~\ref{fig:received_rf} is first separated into its in-phase and quadrature components. These components are then quantized using a predefined number of bits, and the resulting quantized signals are combined back into a complex signal. Consequently, the system does not include controllable analog processing.

Next, a channel-independent spectrogram $\myVec{cs}[z]$ is generated using a modified spectrogram generation function. The spectrogram is computed using a Hann window with a length of $512$ samples and an overlap of $256$ samples. Additionally, the generated spectrogram is cropped to retain the middle $80\%$ of the data, effectively focusing on the most relevant portion of the signal, as illustrated in Fig.~\ref{fig:spectrogram_rf}.

The spectrogram $\myVec{cs}[z]$ serves as the input to a digital \ac{dnn} for feature extraction and classification. The feature extraction is performed by a ResNet-like \acp{cnn}, based on the architecture used in the \cite{LoRa}. This network consists of an initial convolutional layer followed by several residual blocks. The output from the residual blocks is passed through an average pooling layer, flattened, and then fed into a fully connected layer, with a final fully connected layer that maps the features to an $18$-dimensional output, corresponding to the number of transmission modes. This output is then used to recover the correct class of signal. The overall task-based acquisition system is depicted in Fig.~\ref{fig:rf_overview}.

\subsubsection{Results}

\label{ssec:RadioResults}
\begin{figure}
    \centering
    \begin{subfigure}{.45\textwidth}
        \centering
        \includegraphics[width=\linewidth]{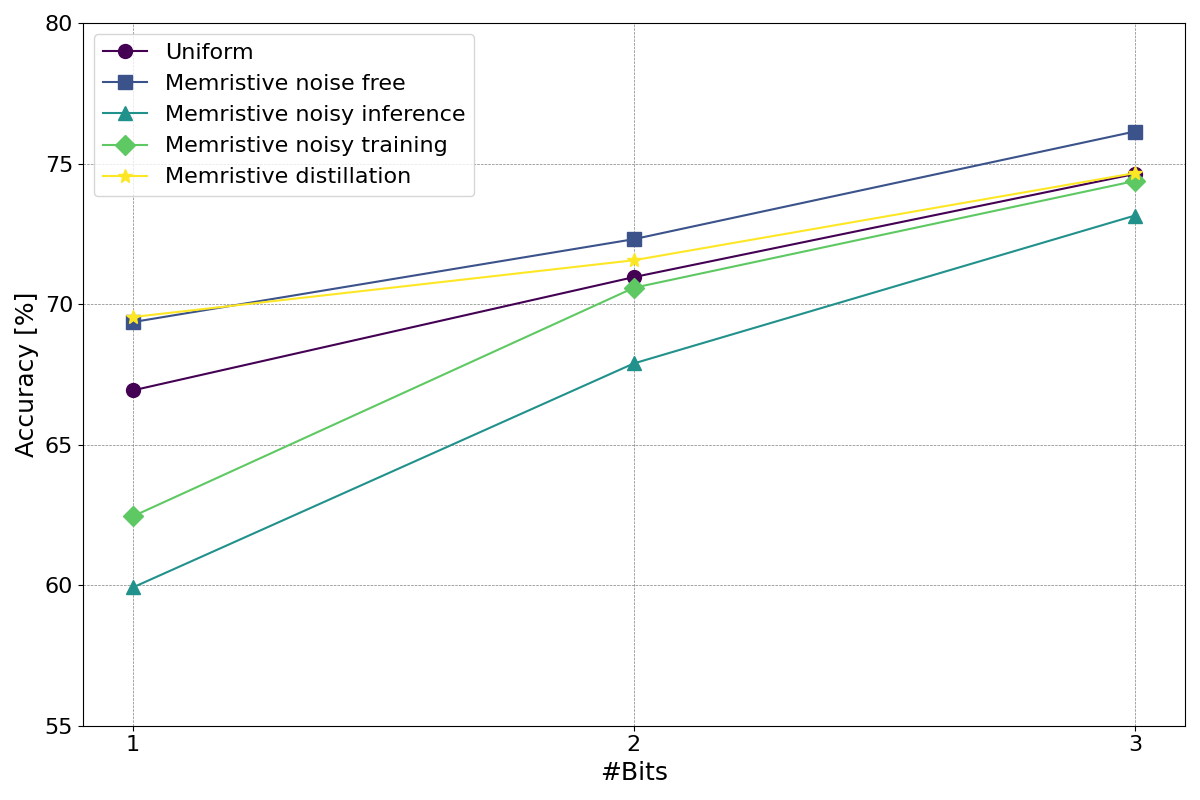}
        \caption{Accuracy}
        \label{fig:accuracy_radio}
    \end{subfigure}

    \vspace{0.4cm}
    
    \begin{subfigure}{.45\textwidth}
        \centering
        \includegraphics[width=\linewidth]{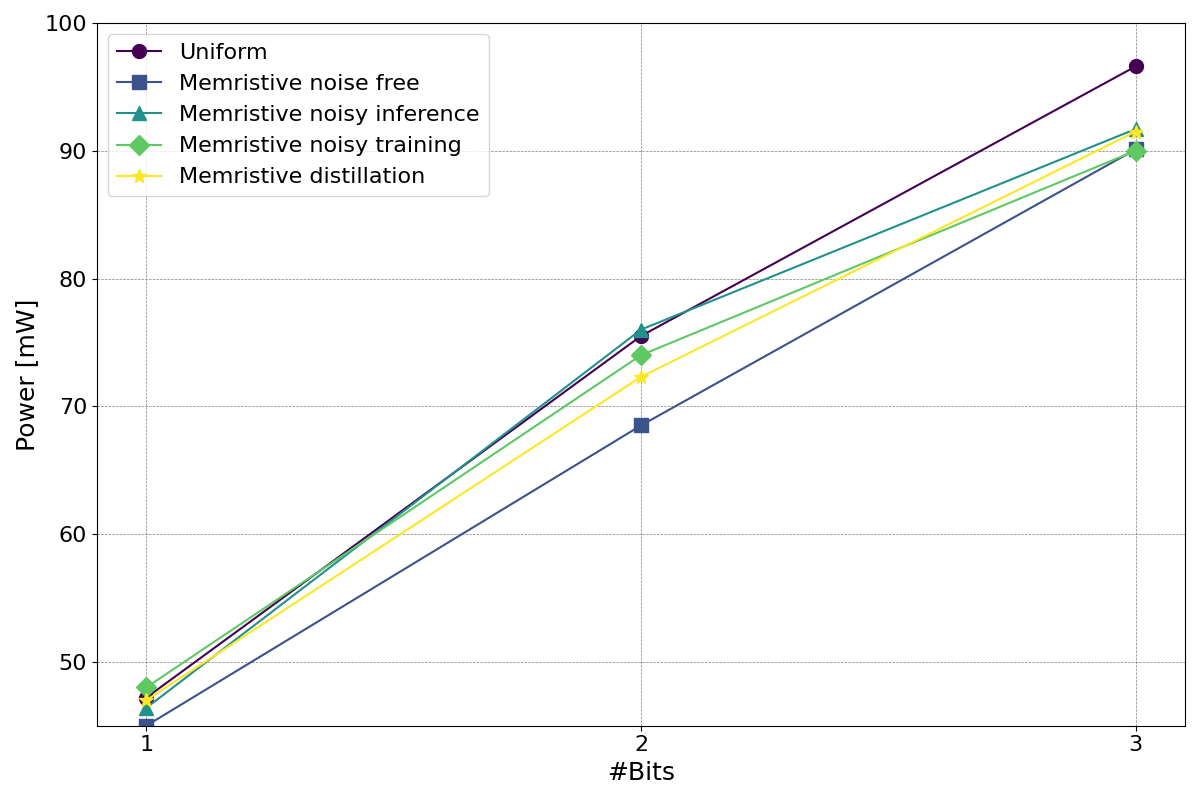}
        \caption{Power}
        \label{fig:power_radio}
    \end{subfigure}
    \vspace{0.4cm}
    \caption{RF signal classification task.} 
    \label{fig:comparison_radio}
\end{figure}

In this study, we evaluate the performance of our task-based memristive ADC network by comparing it with \ac{dnn}-based processing using a uniform ADC across different bit numbers (1, 2, 3). The system employs two ADCs, with one ADC assigned to the I component and another to the Q component. We allowed each memristive ADC to have its own trainable parameters, allowing to learn different decision regions for the I and Q components.

The experimental results, presented in Figs.~\ref{fig:accuracy_radio} and \ref{fig:power_radio}, demonstrate improvements in both power efficiency and accuracy achieved through our approach. Unlike the previous power results shown in Figs.~\ref{fig:power_mnist} and \ref{fig:power_syn}, where power consumption was lower and reported in $\mu$W, the power consumption in the RF dataset is higher. Therefore, we present the results in mW to accurately reflect the performance differences. The memristive noise-free setup exhibits superior performance, achieving higher accuracy and lower power consumption compared to the uniform ADC. This suggests that the utilization of clean memristor-based ADCs significantly enhances system performance.

When introducing noise with variance $\sigma_{i}^{(n)} = 0.03$, the memristive system experiences a slight increase in power consumption and a decrease in accuracy. However, by incorporating noisy training and the distillation method, the system effectively mitigates the impact of noise, ultimately surpassing the uniform ADC in both accuracy and power efficiency.
Our proposed distillation approach  further refines the balance between power consumption and accuracy, achieving the most efficient results among all configurations tested. This underscores the usefulness of our learning method in enhancing the robustness and efficiency of memristive ADC systems, especially in the likely presence of noise induced by the neuromorphic circuitry.

%----------------------------------------------------------------------------------------
%	CONCLUSIONS
%----------------------------------------------------------------------------------------
%\vspace{-0.2cm}
\section{Conclusions}
\label{ssec:conclusions}
%\vspace{-0.15cm} 
We studied power-aware signal acquisition by integrating task-based neuromorphic \acp{adc} within a learning framework. Our method jointly tunes the ADC memristors and associated signal processing to optimize classification performance under power constraints, using a dedicated hardware- and task-aware training method. Simulations across multiple scenarios demonstrated that our approach achieves superior accuracy-power trade-offs compared to traditional uniform ADCs. The task-oriented neuromorphic ADC framework allows improving the performance-power tradeoff in acquisition, even the underlying hardware induces read and write noise. Our results highlight the effectiveness of our approach in real-world, noise-prone applications.
 
%----------------------------------------------------------------------------------------
%	REFERENCES
%----------------------------------------------------------------------------------------
\bibliographystyle{IEEEtran}
\bibliography{IEEEabrv,refs}

\end{document}